\documentclass{article}

\usepackage{graphicx}
    \usepackage[preprint]{neurips_2024}

\usepackage[utf8]{inputenc} 
\usepackage[T1]{fontenc}    
\usepackage{hyperref}       
\usepackage{url}            
\usepackage{booktabs}       
\usepackage{amsfonts}       
\usepackage{nicefrac}       
\usepackage{microtype}      
\usepackage{xcolor}         
\usepackage{amsmath}
\usepackage{graphicx}
\usepackage{subcaption}
\title{VisuLogic: A Benchmark for Evaluating Visual Reasoning in Multi-modal Large Language Models} 
\usepackage{array}
\usepackage{adjustbox}
\usepackage{bbding}
\usepackage{multirow}
\usepackage{wrapfig}
\usepackage{enumitem}
\usepackage{csquotes}
\usepackage{ifsym}
\author{%
\hspace{-1cm}Weiye Xu$^{1,3*\dagger}$,
Jiahao Wang$^{2,3*\dagger}$,
Weiyun Wang$^{3\dagger}$,
Zhe Chen$^{3\dagger}$,
Wengang Zhou$^{1}$,
Aijun Yang$^{2}$,  \\
\textbf{ \hspace{-1cm}Lewei Lu$^{4}$,
Houqiang Li$^{1}$,
Xiaohua Wang$^{2}$,
Xizhou Zhu$^{3}$,
Wenhai Wang$^{3}$,
Jifeng Dai$^{5,3}\textsuperscript{\Envelope}$, 
Jinguo Zhu$^{3}\textsuperscript{\Envelope}$ }\\
  \hspace{-1cm}$^1$University of Science and Technology of China,~~~  $^2$Xi'an Jiaotong University,\\  \hspace{-1cm}$^3$Shanghai Artifcial Intelligence Laboratory,~~~ $^4$SenseTime Research, ~~~
  $^5$Tsinghua University\\
  \hspace{-1cm}\texttt{ustcxwy0271@mail.ustc.edu.cn, wjhwdscience@stu.xjtu.edu.cn,
  lechatelia@gmail.com}
  }

\begin{document}


\renewcommand{\thefootnote}{\fnsymbol{footnote}}
\footnotetext{\textsuperscript{$*$}equal contribution;  \textsuperscript{$\dagger$} interns at OpenGVLab, Shanghai AI Laboratory; \textsuperscript{\Envelope} corresponding author.}
\maketitle
\newcommand{\benchmarkname}{VisuLogic}
\newcommand{\benchmarkshortname}{VisuLogic }
\newcommand{\blue}[1]{\textcolor{blue}{#1}}
\definecolor{reduce-color}{RGB}{67,178,68}
\definecolor{gray-self}{RGB}{116,120,122}
\begin{abstract}

Visual reasoning is a core component of human intelligence and a critical capability for advanced multimodal models. Yet current reasoning evaluations of multimodal large language models (MLLMs) often rely on text descriptions and allow language-based reasoning shortcuts, failing to measure genuine vision-centric reasoning. To address this, we introduce VisuLogic: a benchmark of 1,000 human-verified problems across six  categories (e.g., quantitative shifts, spatial relations, attribute comparisons). These various types of questions can be evaluated  to assess the visual reasoning capabilities of MLLMs from multiple perspectives.  We evaluate leading MLLMs on this benchmark and analyze their results to identify common failure modes. Most models score below 30\% accuracy—only slightly above the 25\% random baseline and far below the 51.4\% achieved by humans—revealing significant gaps in visual reasoning. Furthermore, we provide a supplementary training dataset and a reinforcement-learning baseline to support further progress. Code, data, and baselines are available at \href{https://visulogic-benchmark.github.io/VisuLogic}{https://visulogic-benchmark.github.io/VisuLogic}.
\end{abstract} 

\section{Introduction}
\begin{figure}[h]  
    \centering
    \includegraphics[width=1.0\textwidth]{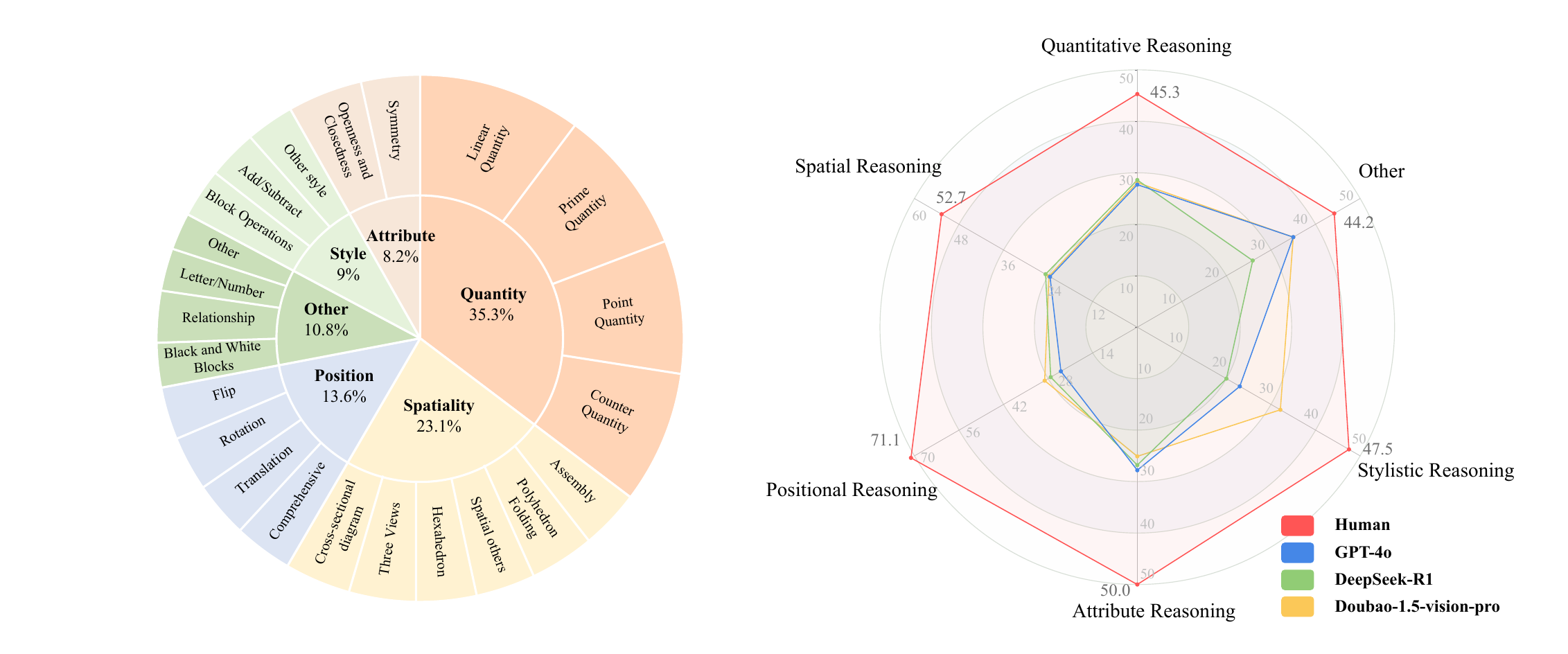}
    \caption{\textbf{Composition of the VisuLogic benchmark and performance of representative MLLMs.} The left figure shows the distribution of the 6 categories and their subcategories in VisuLogic. The right figure shows accuracies (\%) achieved by MLLMs and by human on each category of VisuLogic.}  
    \label{fig:overview}  
    \vspace{-1em}
\end{figure}





Reasoning, as fundamental component of human intelligence, has become a critical criterion in evaluating progress toward Artificial General Intelligence (AGI)~\cite{goertzel2007artificial,wang2024exploring}. Recent advancements in Large Language Models (LLMs) have demonstrated substantial improvements in reasoning capabilities across complex domains such as mathematics~\cite{peng2025lmmr1,yang2024qwen2,xu2024chatglm,meng2025mm}, logical reasoning~\cite{wan2024logicasker,xu2023symbol,feng2023language,liu2023logicot} and coding~\cite{ahmad2025opencodereasoning,jiang2024self,li2025structured,huang2023codecot}. Techniques like Chain-of-Thought (CoT)~\cite{wei2022cot} prompting and test-time compute scaling (e.g., OpenAI o1~\cite{jaech2024openaio1} and Deepseek-R1~\cite{deepseekai2025deepseekr1incentivizingreasoningcapability}) have significantly enhanced the reasoning performance of LLMs~\cite{deepseekai2025deepseekr1incentivizingreasoningcapability, goertzel2007artificial, wang2024exploring}. Along with the rapid development of language reasoning research for LLMs, considerable progress~\cite{yang2025r1onevision, peng2025lmmr1, meng2025mm, chen2025r1v, liu2025visual, wang2025visualprm, liu2025othink, shen2025vlmr1, wang-2025-open-r1-video} has been made in improving multimodal reasoning capability of Multimodal Large Language Models (MLLMs).

\begin{figure}[t]
    \centering
    \begin{subfigure}[b]{1\linewidth}
        \centering
        \includegraphics[width=\linewidth]{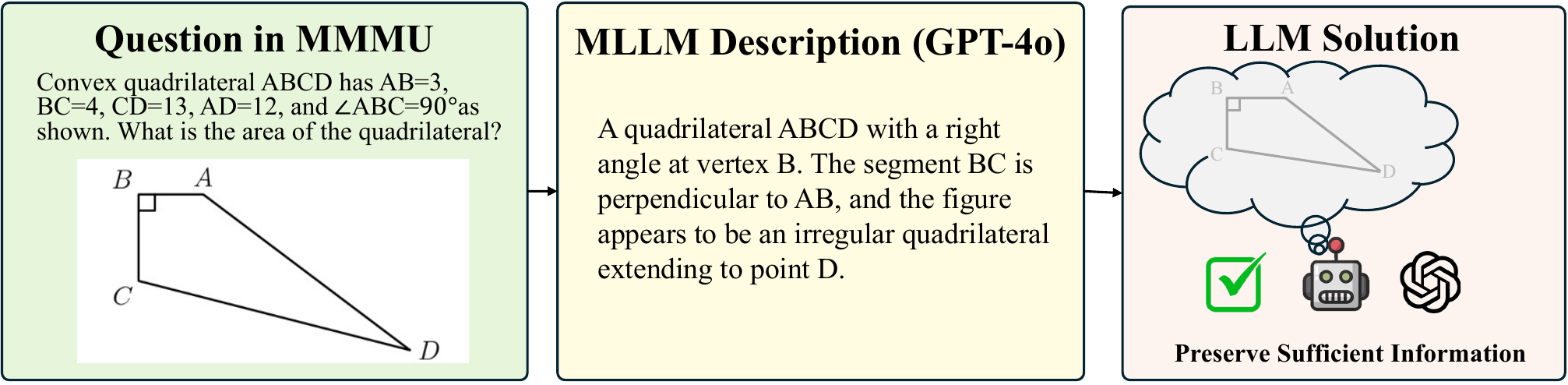}
        \caption{Pipeline of ``MLLM description$\rightarrow$LLM'' for Question in MMMU~\cite{yue2023mmmu}. It is trivial that SOTA MLLMs  extract key visual details, thereby enabling the LLM to  answer questions solely based on language reasoning.}
        \label{fig:aption-examples-subfig1}
    \end{subfigure}
    \hfill
    \begin{subfigure}[b]{1\linewidth}
        \centering
        \includegraphics[width=\linewidth]{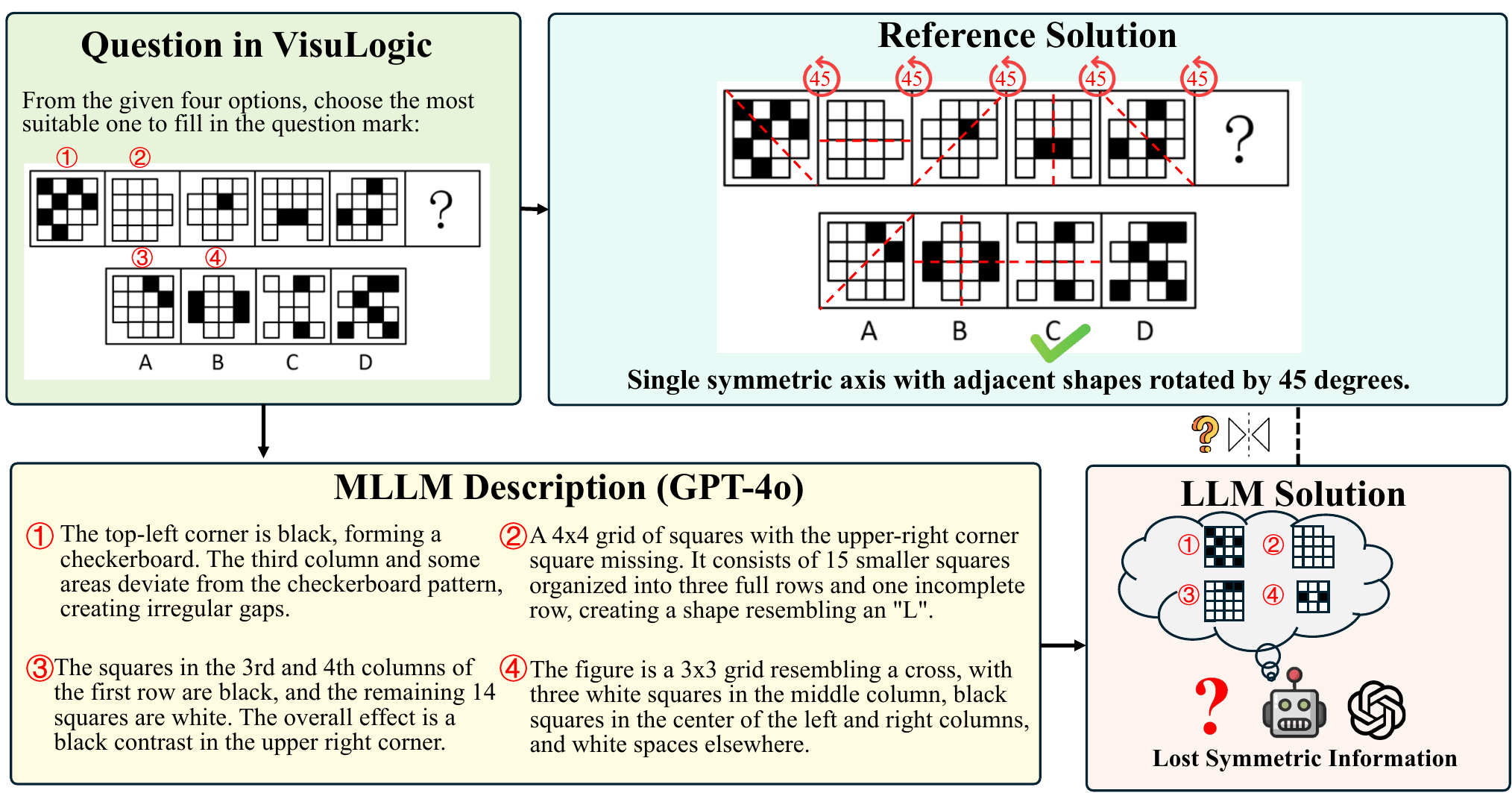}
        \caption{Pipeline of ``MLLM description$\rightarrow$LLM'' for Question in VisuLogic. Even SOTA MLLMs struggle to describe images precisely, leading to ambiguous interpretations.}
        \label{fig:aption-examples-subfig2}
    \end{subfigure}
    \caption{\textbf{Comparison of the ``MLLM description$\rightarrow$LLM'' pipeline on two benchmarks.} In MMMU, detailed descriptions lead to correct solutions, while in  VisuLogic, critical visual cues (\textit{e.g.}, symmetry, rotation) can be easily lost, 
    causing the LLM to misinterpret the image. This highlights that textual reasoning alone is insufficient, underscoring the benchmark’s demand for robust and in-depth visual reasoning.}
    \label{fig:caption-examples}
    \vspace{-0.5em}
\end{figure}

\begin{figure}[t]  
    \centering
    \includegraphics[width=1.0\textwidth]{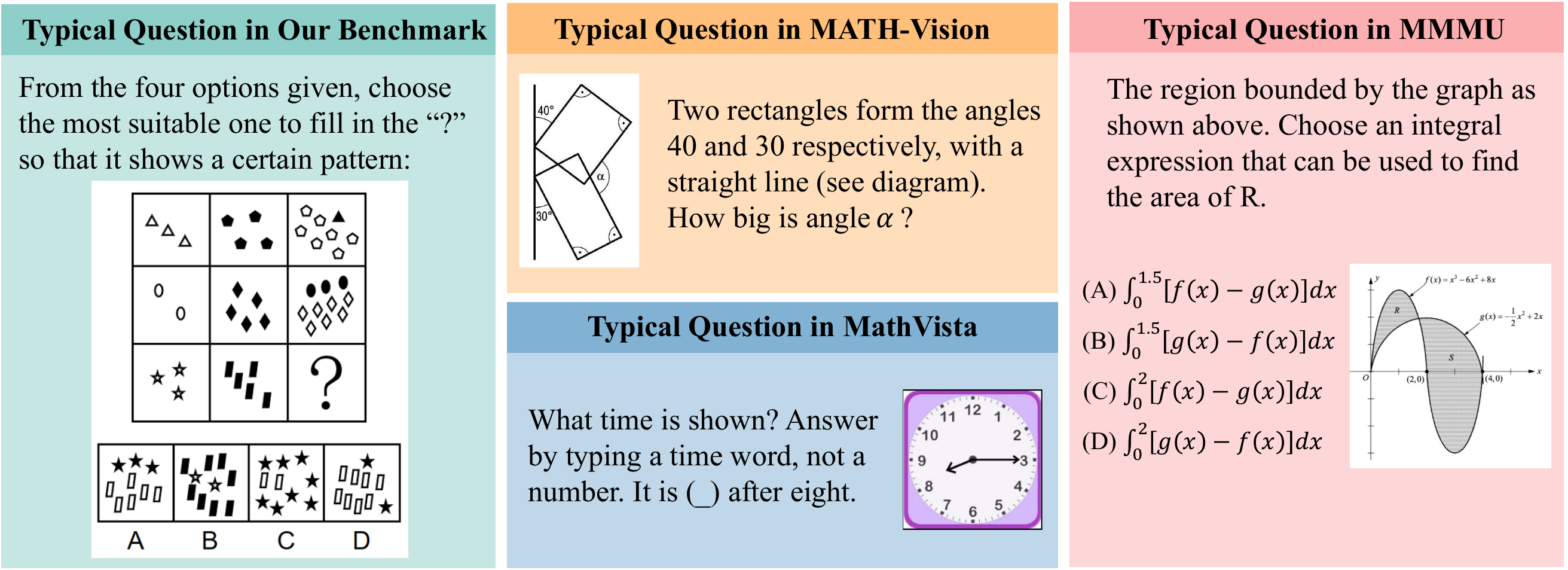}
    \caption{\textbf{Comparison of questions from different Benchmarks.} Compared to MathVista~\cite{lu2023mathvista}, MathVision~\cite{wang2024measuring}, and MMMU~\cite{yue2023mmmu}, VisuLogic focuses more explicitly on assessing pure visual reasoning capabilities.}  
    \label{fig:2_example}  
    \vspace{-0.5em}
\end{figure}

These methods, which often incorporate reinforcement learning techniques~\cite{chen2025r1v, liu2025visual, peng2025lmmr1} to enhance the reasoning capabilities of MLLMs, have achieved some early successes~\cite{yang2025r1onevision,peng2025lmmr1,meng2025mm,chen2025r1v,liu2025visual,liu2025othink,shen2025vlmr1}. However,  they typically rely on existing multi-modal benchmarks that  struggle to accurately capture a model’s core visual reasoning ability. 
For example, VLM-R1~\cite{shen2025vlmr1} assesses “visual reasoning” with referring expression comprehension tasks~\cite{yu2016modeling, mao2016generation,lai2024lisareasoningsegmentationlarge}, yet these tasks primarily focus on object localization, demanding only basic perceptual skills rather than more advanced visual cognitive processes. 
Meanwhile, several works~\cite{meng2025mm,peng2025lmmr1,yang2025r1onevision} adopt mathematical problem-solving benchmarks that include diagrams—such as MathVista~\cite{lu2023mathvista}, MathVerse~\cite{zhang2024mathverse}, and MathVision~\cite{wang2024measuring}—to evaluate visual reasoning. In practice, however, as~\cite{zhang2024mathverse} observes, many MLLMs translate these visual clues into
 textual descriptions and then rely on standard language reasoning. 
  This approach can incorrectly attribute language-driven results to visual reasoning, resulting in a misleading assessment of the model's visual reasoning capabilities~\cite{zhang2024mathverse,hao2025can}. 
Consequently, designing new benchmarks that explicitly focus on vision-centric reasoning—rather than conflating it with text-based reasoning—remains critical for advancing MLLMs’ visual reasoning capacities.

\begin{figure}[t]  
    \centering
    \includegraphics[width=1.0\textwidth]{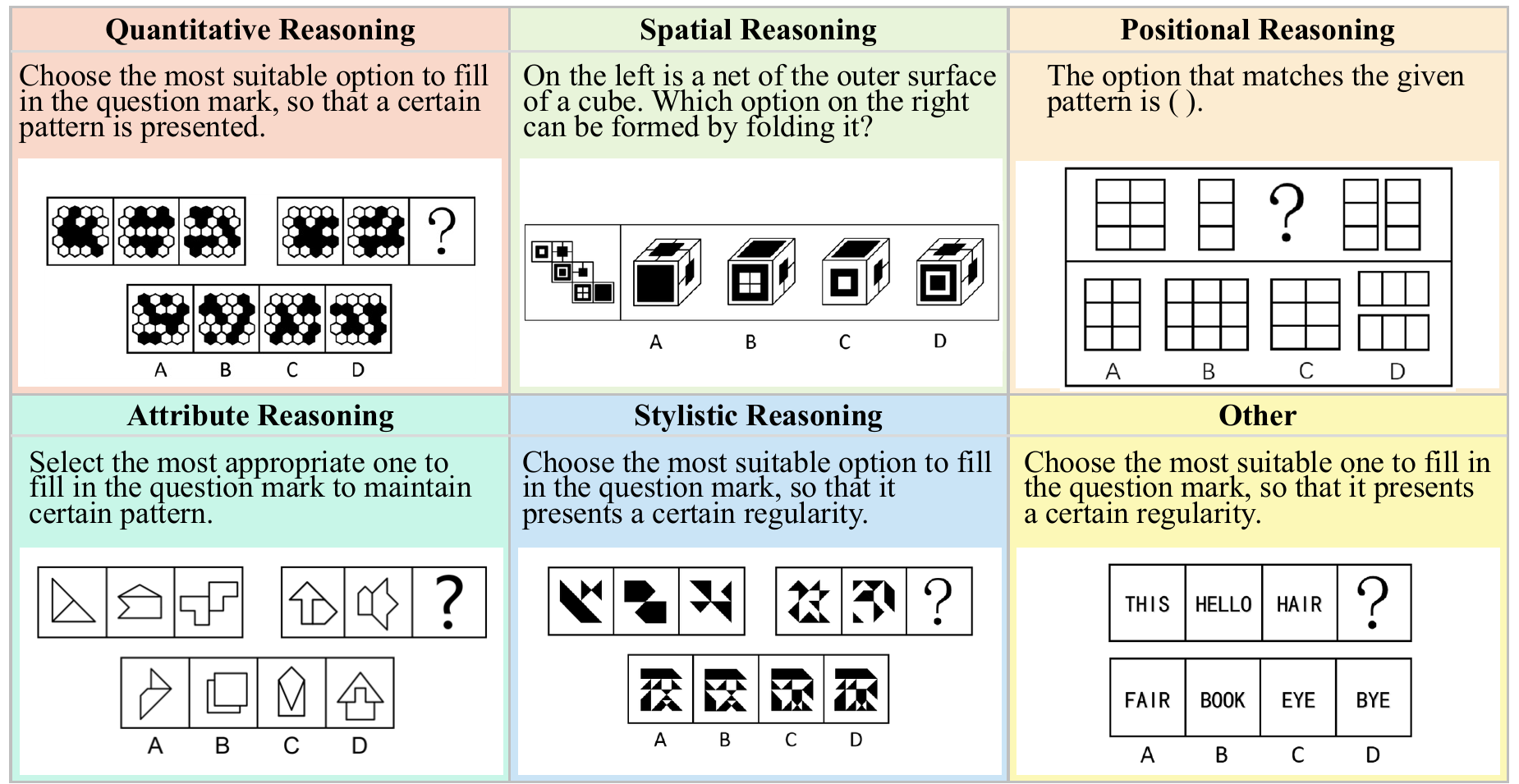}
    \caption{\textbf{Question examples of different categories in our VisuLogic Benchmark.} VisuLogic contains 6 categories of questions, which require models' abilities in visual logic reasoning.}  
    \label{fig:4_problem_example}  
\end{figure}

To address this limitation, we propose VisuLogic, a novel benchmark specifically designed to evaluate  visual reasoning abilities in multimodal models without mixing them with purely text-based reasoning (see Figure~\ref{fig:2_example}). VisuLogic comprises carefully constructed tasks that span multiple reasoning categories (see Figure~\ref{fig:overview}). As shown in Figure~\ref{fig:4_problem_example}, these tasks are classified into six key types, such as Quantitative Reasoning, which requires understanding and deducing shifts in the quantity of certain elements within an image.
In contrast to existing benchmarks, as demonstrated in Figure~\ref{fig:caption-examples}, state-of-the-art (SOTA) MLLMs often omit crucial visual details when describing VisuLogic problems, making it difficult for them to rely solely on a text-based inference shortcut. Indeed, even humans would find it challenging to capture every essential visual cue in a single description, so effectively tackling VisuLogic demands more robust, vision-centric reasoning. By reducing reliance on textual inference shortcuts, VisuLogic thus provides a stringent evaluation of MLLMs’ genuine visual reasoning capabilities.

We conducted a comprehensive evaluation and systematic analysis to assess current models’ visual reasoning capabilities. When leading text-only LLMs were supplied with detailed descriptions in place of raw images, their accuracy—Doubao-1.5-Pro (26.6\%), Claude-3.7-Sonnet (25.9\%) and Qwen2.5-72B-Instruct~\cite{qwen2.5} (28.0\%)—barely exceeded the random-chance baseline of 24.9\%. This clearly demonstrates that textual reasoning alone are insufficient for solving our VisuLogic tasks. Even state-of-the-art multimodal arge language models (MLLMs)—including GPT-4o~\cite{hurst2024gpt}, Doubao-1.5-Vision-Pro, Gemini-2.0-Pro-Exp~\cite{team2023gemini} and InternVL3-78B~\cite{zhu2025internvl3exploringadvancedtraining}—achieve only 26.3\%, 28.1\%, 28.0\% and 27.7\%, respectively, whereas human participants reached 51.4\%. The substantial gap between these results and human performance underscores the challenge of robust visual reasoning in current MLLMs.
Furthermore, we applied a simple reinforcement-learning (RL) fine-tuning step on our supplementary training dataset: this boosted the baseline model’s accuracy from 25.5\% to 31.1\%, outperforming both open-source and closed-source counterparts. These findings illustrate the promise of the RL technique for strengthening MLLMs’ visual reasoning capabilities.

In summary, our contributions are as follows: 
\begin{itemize}[leftmargin=1em]
\vspace{-0.7em}
\item We propose a challenging visual reasoning benchmark that is inherently difficult to articulate using language, providing a more rigorous evaluation of the visual reasoning capabilities of MLLMs.
\vspace{-0.4em}
\item We conduct comprehensive experiments to evaluate and analyze the benchmark, including extensive evaluations and comparative studies of various MLLMs under different setting. 
\vspace{-0.4em}
\item We identify the  RL technique as a promising direction for improving the visual reasoning capabilities of MLLMs. Furthermore, we release both the training code and data to facilitate future research.
\end{itemize}

\section{Related Work}


\noindent\textbf{Multi-modal Large Language Models.} Recent years have witnessed substantial advancements in Multi-modal Large Language Models (MLLMs). Early works like BLIP~\cite{li2022blip, li2023blip2} and Flamingo~\cite{alayrac2022flamingo} introduce lightweight parameters between vision transformer~\cite{dosovitskiy2020vit} (ViT) and LLMs, laying the groundwork for multimodal perception. Subsequent efforts, such as LLaVA~\cite{llava} and MiniGPT-4~\cite{zhu2023minigpt}, integrate instruction tuning, further enhancing the performance of MLLMs.  Proprietary models like GPT-4o~\cite{hurst2024gpt} and Gemini-Pro~\cite{team2023gemini} have advanced MLLM performance on complex multimodal tasks, while open-source models such as Qwen-VL series~\cite{Qwen-VL,wang2024qwen2,Qwen2.5-VL} and InternVL series~\cite{chen2024far,chen2024internvl,gao2024mini,chen2024expanding,zhu2025internvl3exploringadvancedtraining} achieve competitive results through optimized architectural design, dataset expansion and training paradigm improvements. Meanwhile, some related studies further advance the ability of large models by incorporating new modalities (e.g., audio~\cite{fang2024llama, defossez2024moshi, xie2408miniomni}, point clouds~\cite{guo2023point, chen2023pointgpt}, video~\cite{zhao2023antgpt, chen2023vast}) and by supporting more tasks  (e.g., grounding~\cite{xu2024vlm-grounder, wang2025learningvisualground}, computer usage~\cite{niu2024screenagent,bai2025digi}). Notably, limited research attempts to enhance the reasoning capabilities of MLLMs. Some pioneering works, such as R1-Onevision~\cite{yang2025r1onevision}, LMM-R1~\cite{peng2025lmmr1}, MM-EUREKA~\cite{meng2025mm}, R1-V~\cite{chen2025r1v}, Visual-rft~\cite{liu2025visual}, Visualprm~\cite{wang2025visualprm}, OThink-MR1~\cite{liu2025othink}, VLM-R1~\cite{shen2025vlmr1}, and Open-r1-Video~\cite{wang-2025-open-r1-video} have explored the visual reasoning capabilities of MLLMs through Reinforcement Learning (RL), but they are still in the nascent stage.

\noindent\textbf{Multimodal Benchmarks.} With the development of MLLMs, multimodal benchmarks have also evolved significantly~\cite{li2024surveybenchmarksmultimodallarge}.  Early benchmarks primarily address visual perception tasks through simple tasks like visual question answering (VQA)~\cite{chen2015microsoftcococaptionsdata, lin2015microsoftcococommonobjects, kay2017kineticshumanactionvideo, xu2017video}, image captioning~\cite{nguyen2023improving,dong2024benchmarking,ke2019reflective} and referring expression comprehension~\cite{yu2016modeling, mao2016generation}. Subsequent works expand the capability coverage of benchmarks into more specialized domains: OCRBench~\cite{liu2024ocrbench}, Chartqa~\cite{masry2022chartqa} and DocVQA~\cite{mathew2021docvqa} assess textual content extraction; AgentBench~\cite{liu2023agentbench} and ToolEyes~\cite{ye2024tooleyes} test tool usage capabilities; and egocentric perception benchmarks~\cite{mangalam2023egoschema, cheng2024egothink} quantify first-person scene interpretation. Despite the progress, they ignore the evaluation of visual reasoning abilities~\cite{zhang2019raven,yue2023mmmu}. Recently, some benchmarks have made explorations in examining MLLMs' visual reasoning abilities, but methodological deficiencies still cause limitations to assess the intrinsic visual reasoning capabilities~\cite{hao2025can, akter2024visreascomplexvisualreasoning, xiao2024logicvista}. InfiMM-Eval~\cite{han2023infimm} test reasoning abilities around daily life, lacking deep-level reasoning scenarios. MMMU~\cite{yue2023mmmu} and Emma~\cite{hao2025can} provide benchmarks demanding advanced reasoning abilities in fields such as chemistry and physics, but they ignore questions around the images' fundamental visual components (e.g., shapes, elements). While mathematical benchmarks~\cite{wang2024measuring, lu2023mathvista, he2024olympiadbench, qiao2024we, zhang2024mathverse, gupta2024polymathchallengingmultimodalmathematical} evaluate mathematical reasoning with geometric and diagram problems included, they focus on math capabilities but disregard logical analysis about the vision information. LogicVista~\cite{xiao2024logicvista} provides a multimodal logical reasoning benchmark, its visual questions lack analytical depth—dominated by single-hop, superficial queries in limited data scope. Unlike previous works, we introduce a challenging benchmark focused specifically on the domain of visual logical reasoning.

\section{VisuLogic}



In this section, we first describe the VisuLogic data-curation pipeline, which comprises three key stages: data collection, quality control, and the detailed taxonomy. We then report the benchmark's construction statistics, including total size, answer-option distributions, and category-level proportions. Finally, we introduce a supplementary training dataset—consisting of questions analogous to those in VisuLogic—designed to bolster future research and facilitate community engagement.

\subsection{Data Curation Pipeline}

\begin{figure}[tbp]  
    \centering
    \includegraphics[width=1.0\textwidth]{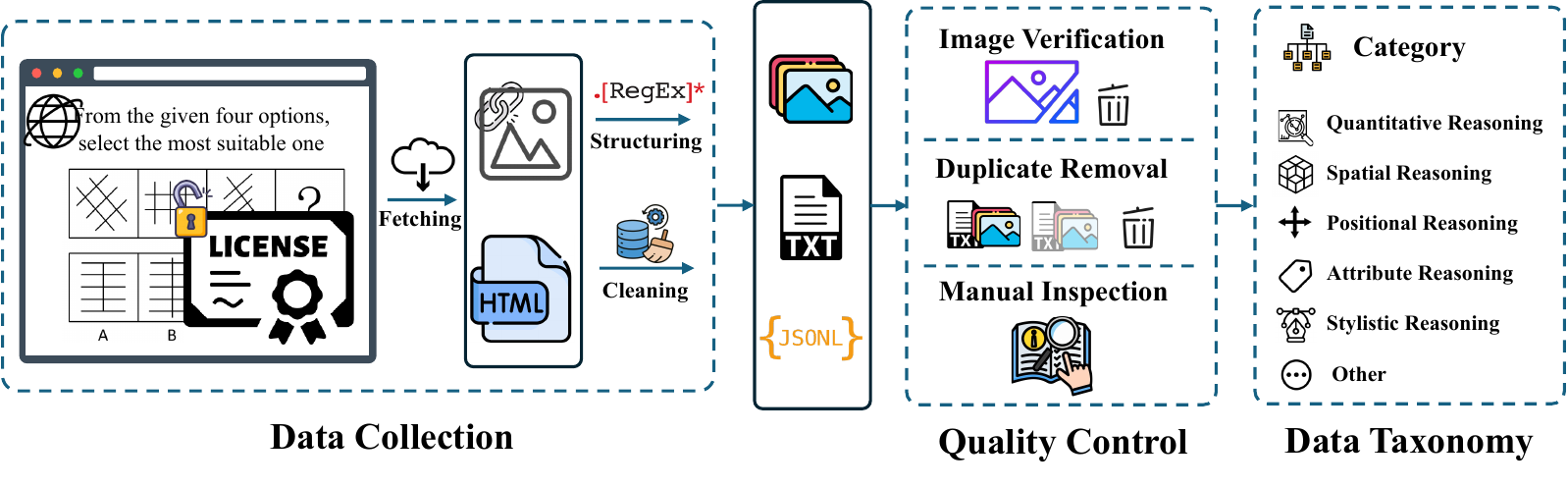}
    \vspace{-1em}
    \caption{\textbf{Data curation pipeline of VisuLogic.} The pipeline includes Data Collection, Quality Control and Data Taxonomy.}  
    \label{fig:3_pipeline}  
    \vspace{-1em}
\end{figure}

\noindent\textbf{Data Collection.} We construct the VisuLogic dataset by sourcing all questions from publicly available online resources in compliance with relevant licenses and regulations. As shown in Figure~\ref{fig:3_pipeline},  our automated data processing pipeline comprises three stages: 1) \textbf{Fetching}: We employ Playwright~\footnote[1]{https://github.com/microsoft/playwright} to systematically scrape raw web content, supplemented by custom parsing scripts that extract question–answer pairs. 2) \textbf{Cleaning}: We remove noise, irrelevant content, and extraneous HTML markup (\emph{e.g.}, \texttt{<div>}) to ensure the integrity of the textual data. 3) \textbf{Structuring}: We standardize the cleaned text and images by structuring all information (such as question text, metadata) in JSON Lines (JSONL) format.

\vspace{0.5em}

\label{sec:qualitycontrol}
\noindent\textbf{Quality Control.} To ensure the reliability of the benchmark dataset, we employ a three-stage data validation procedure: 1)~\textbf{Image Verification}: Each image referenced in the questions is checked for existence and correct formatting; any item that fails to meet the criteria is removed following human review. 2)~\textbf{Duplicate Removal}: We eliminate redundant entries at both the text and image levels by (i) detecting lexical overlap among text segments and (ii) applying perceptual hashing (pHash) to identify visually similar images. 3)~\textbf{Manual Checking}: After automated filtering, we perform a thorough human-led review of every remaining entry to confirm its validity and ensure dataset reliability.

\vspace{0.5em}

\noindent\textbf{Data Taxonomy.} We categorize all collected data into a taxonomy of six primary classes based on expert human annotation of the reasoning skills each question requires. Annotators first tag questions according to the targeted reasoning competency; these annotated tags are then analyzed and merged into five primary categories. A subsequent human review ensures that every question is accurately classified, with any ambiguous instances consolidated under the ``Other'' category. 
Specifically, we define each category as follows. 
\textbf{Quantitative Reasoning} focuses on changes in the number or count of graphical elements (for example, points, lines and angles) and on arithmetic relationships among shapes. \textbf{Spatial Reasoning} requires mentally reconstructing three-dimensional shapes from two-dimensional figures, folding or unfolding surfaces, and integrating three-dimensional structures. \textbf{Positional Reasoning} examines transformations such as translation, rotation and reflection of objects while preserving their fundamental elements. \textbf{Attribute Reasoning} involves intrinsic properties of shapes, including symmetry (axial or central), curvature and measures of openness or closedness. \textbf{Stylistic Reasoning} entails alterations in stylistic features such as overlay, subtraction and assessments of shape similarity or difference. \textbf{Other} encompasses questions that fall outside the preceding categories, including those involving letters, alphanumeric symbols or other specialized characters.

\subsection{Dataset Statistics}
\label{sec:Dataset-Statistics}

Following data curation and validation, VisuLogic comprises 1,000 single-choice questions.  Figure~\ref{fig:overview} (left) illustrates the category distribution: Quantitative Reasoning (35.3\%), Spatial Reasoning (23.1\%), Positional Reasoning (13.6\%), Attribute Reasoning (8.2\%), Stylistic Reasoning (9.0\%), and Other (10.8\%). Correct answer options are evenly balanced, with the proportions distributed as follows:  A (23.1\%), B ( 26.7\%), C (25.2\%), and D (25.0\%).

\subsection{Supplementary Training Dataset}
To facilitate further investigation of visual reasoning, we provide an auxiliary training set of 4,296 question–answer pairs drawn from the same domains and subjected to identical validation procedures to prevent overlap with the benchmark. The training split mirrors the primary taxonomy, with category proportions of Quantitative Reasoning (30.7\%), Spatial Reasoning (25.5\%), Positional Reasoning (13.0\%), Attribute Reasoning (8.8\%), Stylistic Reasoning (9.9\%), and Other (12.1\%). 





\begin{figure}[h]  
    \centering
    \includegraphics[width=1.0\textwidth]{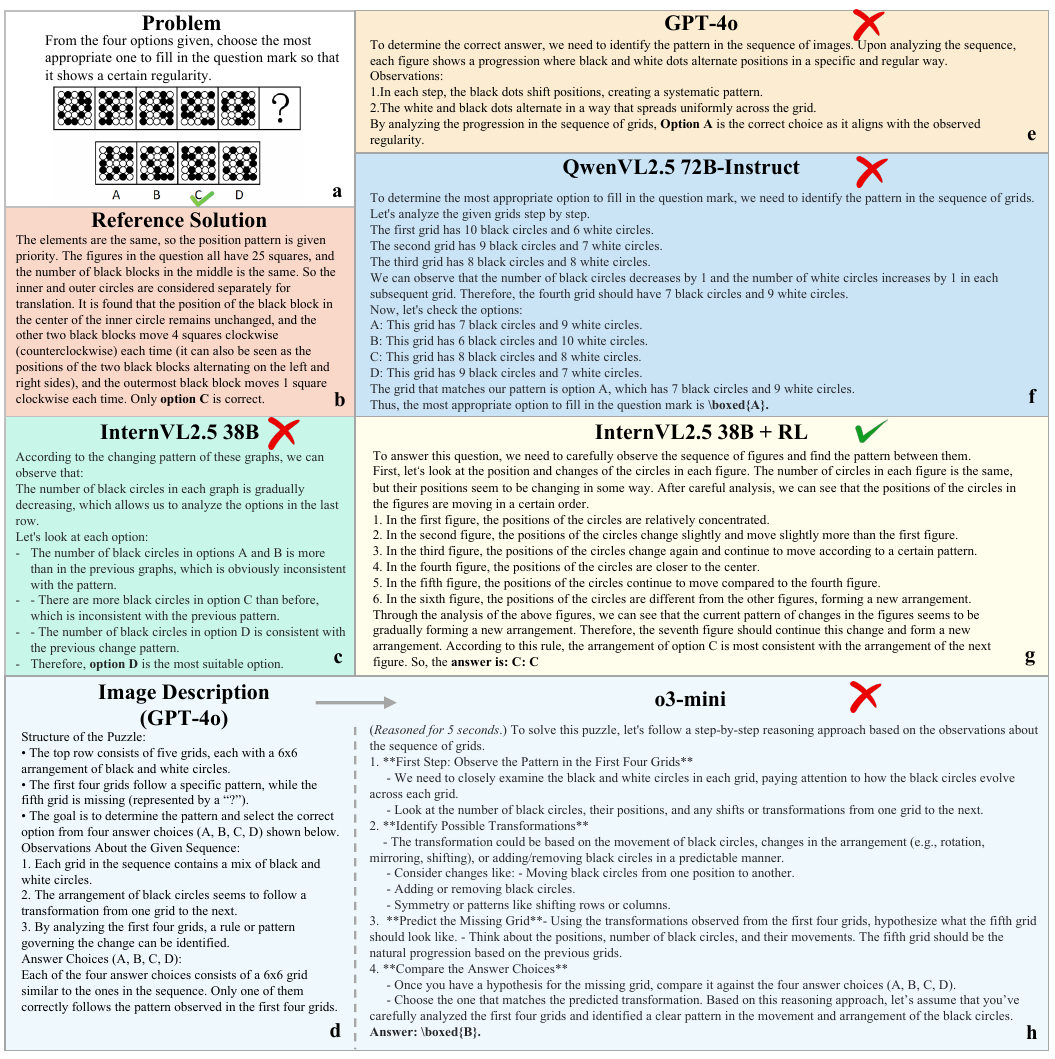}
    \caption{\textbf{Solution examples generated by different models.} Reference solution and outputs generated by GPT-4o~\cite{hurst2024gpt}, Qwen2.5VL-72B-Instruct~\cite{Qwen2.5-VL}, InternVL2.5-38B~\cite{chen2024internvl}, and InternVL2.5-38B with RL. Additionally, the image description and solution from LLMs (o3-mini) are also illustrated.}  
    \label{fig:5_model_output}  
    \vspace{-0.5em}
\end{figure}
\begin{table}[ht]
  \tiny
  \caption{\textbf{Cross-Modal performance with CoT prompts on VisuLogic.} The table shows the evaluation scores of baseline references, LLMs, and MLLMs, which illustrates a gap between humans' and models' capabilities. Top performers per category are \textbf{bolded}, with secondary leaders \underline{underlined}.}
 \label{sample-table-8x8}
  \centering
    \resizebox{\linewidth}{!}{
  \begin{tabular}{@{}l|c|cccccc@{}}
    \toprule
    \textbf{Models} & \textbf{Overall} & \textbf{Quantity} & \textbf{Spatiality} & \textbf{Position} & \textbf{Attribute} & \textbf{Style} & \textbf{Other} \\
    \midrule
    \multicolumn{8}{c}{\tiny{\textbf{References}}} \\
    \midrule
    Human & 51.4 & 45.3 & 52.7 & 71.1 & 50.0 & 47.5 & 44.2 \\
    Random  & 24.9 & 25.7 & 25.4 & 22.7 & 23.4 & 24.3 & 26.1 \\
    \midrule
    \multicolumn{8}{c}{\tiny{\textbf{Open Source LLM (MLLM Description$\rightarrow$LLM)}}} \\
    \midrule
     Deepseek-R1~\cite{deepseekai2025deepseekr1incentivizingreasoningcapability} & \underline{26.6} & \underline{27.7} & \underline{23.5} & 24.0 & \textbf{27.8} & \underline{23.0} & \textbf{35.0} \\
    Qwen2.5-72B-Instruct~\cite{qwen2.5} & \textbf{28.0} & \textbf{30.2} & \textbf{24.4} & \textbf{27.5} & \underline{26.5} & \textbf{26.8} & \underline{30.8} \\
    QwQ-32B~\cite{qwq32b} &  22.8 & 24.6 & 20.1 & \underline{25.4} & 19.0 & 20.7 & 24.0 \\
    \midrule
    \multicolumn{8}{c}{\tiny{\textbf{Close Source LLM (MLLM Description$\rightarrow$LLM)}}} \\
    \midrule
    GPT-4 (20240613)~\cite{achiam2023gpt} & 23.6 & 21.2 & \underline{22.5} & 21.3 & \underline{25.6} & 23.3 & \textbf{35.2} \\
    o3-mini (20250131) & 24.6 & \underline{27.8} & 18.8 & 24.5 & 21.7 & \underline{25.6} & 28.4 \\
    Gemini-2.0-Flash-Thinking (20250121)~\cite{team2023gemini} & 23.4 & 23.2 & \textbf{26.0} & 16.9 & 17.1 & 21.1 & \underline{33.3} \\
    Claude-3.7-Sonnet (20250219) & \underline{25.9} & 26.6 & \underline{22.5} & \textbf{25.0} & \textbf{28.0} & \underline{25.6} & 30.6 \\
    Doubao-1.5-Pro-32k (20250115) & \textbf{26.6} & \textbf{30.0} & \underline{22.5} & \textbf{25.0} & \underline{25.6} & \textbf{30.0} & 24.1 \\
    \midrule
    \multicolumn{8}{c}{\tiny{\textbf{Close Source MLLMs}}} \\
    \midrule
    GPT-4o-mini (20240718) & 24.3 & 27.2 & 23.4 & 23.5 & 18.3 & \underline{31.1} & 16.7 \\
    GPT-4o (20240806)~\cite{hurst2024gpt} & 26.3 & \underline{28.6} & 24.7 & 27.2 & 26.8 & 20.0 & 25.9 \\
    Kimi-latest ~\cite{team2025kimi} & 25.9 & 24.9 & \textbf{29.4} & 26.5 & \underline{28.0} & 16.7 & 26.9 \\
    Doubao-1.5-Vision-Pro-32k (20250115)  & \textbf{28.1} & 28.1 & 23.8 & \textbf{29.1} & 25.1 & \textbf{32.1} & \textbf{35.0} \\
    Gemini-2.0-Pro (20250205)~\cite{team2023gemini} & \underline{28.0} & \textbf{29.7} & 24.2 & \underline{27.9} & \textbf{30.5} & 22.2 & \underline{33.3} \\
    Claude-3.7-Sonnet (20250219) & 24.8 & 22.7 & \underline{27.3} & \underline{27.9} & \underline{28.0} & 22.2 & 22.2 \\
    \midrule
    \multicolumn{8}{c}{\tiny{\textbf{Open Source MLLMs}}} \\
    \midrule
    LLaVA-v1.5-7B~\cite{liu2023llava} & 24.6 & 26.1 & 24.2 & 23.5 & 17.1 & \underline{31.1} & 22.2 \\
    LLaVA-OneVision-7B (SI)~\cite{li2024llavaonevisioneasyvisualtask} & 25.3 & 22.4 & \underline{27.3} & \textbf{33.1} & 23.2 & 25.6 & 22.2 \\
    ShareGPT4V~\cite{chen2023sharegpt4v} & 23.4 & 24.9 & 22.1 & 23.5 & 19.5 & 28.9 & 19.4 \\
    MiniCPM-o-2.6~\cite{yao2024minicpm} & 25.3 & 25.6 & 23.0 & 27.3 & 21.9 & 24.5 & 29.9 \\
    GLM-4v-9B~\cite{glm2024chatglm} & 24.3 & 22.4 & 23.7 & 28.3 & 26.0 & 24.1 & 25.3 \\
    Ovis2-8B~\cite{lu2024ovis} & 25.6 & 26.1 & 23.8 & 27.2 & \textbf{28.0} & 25.6 & 24.1 \\
    mPLUG-Owl3-7B-241101~\cite{ye2024mplugowl3longimagesequenceunderstanding} & 18.9 & 21.5 & 15.2 & 16.2 & 20.7 & 18.9 & 20.4 \\
    Qwen2.5-VL-7B-Instruct~\cite{Qwen2.5-VL} & 26.0 & 27.6 & 20.9 & 25.2 & 23.2 & \textbf{37.8} & 25.0 \\
    Qwen2.5VL-72B-Instruct~\cite{Qwen2.5-VL} & 26.2 & 25.2 & 23.8 & 27.2 & 25.6 & 25.6 & \textbf{34.3} \\
    QvQ-72B-Preview ~\cite{qvq-72b-preview} & 23.0 & 24.2 & 17.0 & 24.4 & 21.0 & 24.4 & 30.6 \\
    InternVL2.5-38B~\cite{chen2024expanding} & 25.5 & 24.4 & 26.4 & 27.2 & 23.2 & 25.6 & 26.9 \\
    InternVL2.5-78B~\cite{chen2024expanding} & \underline{27.3} & 26.6 & 26.0 & 26.5 & \underline{26.8} & \underline{31.1} & 30.6 \\ 
    InternVL3-38B~\cite{zhu2025internvl3exploringadvancedtraining} & 27.1 & \textbf{28.7} & \textbf{27.6} & 26.1 & 21.4 & 23.9 & 28.5 \\
    InternVL3-78B~\cite{zhu2025internvl3exploringadvancedtraining} & \textbf{27.7} & \underline{27.7} & 26.1 & \underline{31.6} & 26.3 & 21.3 & \underline{32.3} \\
    \midrule
    Qwen2.5-VL-7B-Instruct-SFT  & 25.5 & 24.4 & 26.4 & \underline{27.2} & 23.2 & 25.6 & 26.9 \\
    Qwen2.5-VL-7B-Instruct-RL & \underline{28.0} & 26.6 & \textbf{33.8} & \textbf{29.4} & 23.2 & 18.9 & \underline{29.6} \\
    InternVL2.5-38B-SFT & 27.9 & \underline{30.6} & 29.4 & 20.6 & \underline{25.6} & \textbf{30.0} & 25.0 \\
    InternVL2.5-38B-RL  & \textbf{31.1} & \textbf{31.2} & \underline{31.2} & 26.5 & \textbf{30.5} & \textbf{30.0} & \textbf{38.9}\\
    \bottomrule
  \end{tabular}}
\end{table}

\begin{table}[ht]
  \caption{\textbf{Influence of Chain-of-Thought on model performance.} Positive value changes are highlighted in \textcolor{red}{red}, negative changes in \textcolor{reduce-color}{green}, and statistically insignificant variations (delta < 1\%) are denoted in \textcolor{gray-self}{gray}. With CoT prompts, MLLMs only exhibit tiny improvements in visual reasoning.}
  \tiny
  \label{cot_table}
  \centering
    \resizebox{\linewidth}{!}{
  \begin{tabular}{lllllllll}
    \toprule
    \textbf{Models} & \textbf{CoT} & \textbf{Overall} & \textbf{Quantity} & \textbf{Spatiality} & \textbf{Position} & \textbf{Attribute} & \textbf{Style} & \textbf{Other}  \\
    \midrule
    \midrule
   \multirow{2}{*}{GPT-4o (20240806)} &\Checkmark  & 26.3 & 28.6 & 24.7 & 27.2 & 26.8 & 20.0 & 25.9 \\
    &\XSolid & $26.0_{\textcolor{gray-self}{(-0.3)}}$ & $26.9_{\textcolor{reduce-color}{(-1.7)}}$ & $24.2_{\textcolor{gray-self}{(-0.5)}}$ & $26.5_{\textcolor{gray-self}{(-0.7)}}$ & $23.2_{\textcolor{reduce-color}{(-3.6)}}$ & $24.0_{\textcolor{red}{(+4.0)}}$ & $29.6_{\textcolor{red}{(+3.7)}}$ \\
     \midrule
   \multirow{2}{*}{Kimi-latest} &\Checkmark & 25.9 & 24.9 & 29.4 & 26.5 & 28.0 & 16.7 & 26.9 \\
    &\XSolid & $25.1_{\textcolor{gray-self}{(-0.8)}}$ & $22.9_{\textcolor{reduce-color}{(-2.0)}}$ & $22.5_{\textcolor{reduce-color}{(-6.9)}}$ & $25.0_{\textcolor{reduce-color}{(-1.5)}}$ & $19.5_{\textcolor{reduce-color}{(-7.5)}}$ & $35.6_{\textcolor{red}{(+18.9)}}$ & $24.1_{\textcolor{reduce-color}{(-2.8)}}$ \\
     \midrule
   \multirow{2}{*}{GPT-4o-mini (20240718)} &\Checkmark & 24.3 & 27.2 & 23.4 & 23.5 & 18.3 & 31.1 & 16.7 \\
    &\XSolid & $23.1_{\textcolor{reduce-color}{(-1.2)}}$ & $23.8_{\textcolor{reduce-color}{(-3.4)}}$ & $22.9_{\textcolor{gray-self}{(-0.5)}}$ & $24.3_{\textcolor{gray-self}{(+0.8)}}$ & $17.1_{\textcolor{reduce-color}{(-1.2)}}$ & $30.0_{\textcolor{reduce-color}{(-1.1)}}$ & $18.5_{\textcolor{red}{(+1.8)}}$ \\
     \midrule
   \multirow{2}{*}{Qwen2.5-VL-Instruct-7B} &\Checkmark & 26.0 & 27.6 & 20.9 & 25.2 & 23.2 & 37.8 & 25.0 \\
    &\XSolid  & $25.9_{\textcolor{gray-self}{(-0.1)}}$ & $25.5_{\textcolor{reduce-color}{(-2.1)}}$ & $22.8_{\textcolor{red}{(+1.9)}}$ & $26.4_{\textcolor{red}{(+1.2)}}$ & $25.3_{\textcolor{red}{(+2.1)}}$ & $20.6_{\textcolor{reduce-color}{(-17.2)}}$ & $38.2_{\textcolor{red}{(+13.2)}}$ \\
     \midrule
   \multirow{2}{*}{InternVL2.5-38B} &\Checkmark & 24.9 & 24.1 & 26.4 & 27.2 & 23.2 & 25.6 & 22.2 \\
    &\XSolid & $25.0_{\textcolor{gray-self}{(+0.1)}}$ & $24.6_{\textcolor{gray-self}{(+0.5)}}$ & $25.5_{\textcolor{gray-self}{(-0.9)}}$ & $22.1_{\textcolor{reduce-color}{(-5.1)}}$ & $22.0_{\textcolor{reduce-color}{(-1.2)}}$ & $26.7_{\textcolor{red}{(+1.1)}}$ & $29.6_{\textcolor{red}{(+7.4)}}$ \\
    \bottomrule
  \end{tabular}}
\end{table}

\begin{table}[ht]
  \caption{\textbf{Influence of hint prompts on model performance.} MLLMs exhibit measurable performance enhancements with hint integration, yet retain significant gaps against human performance. In comparison, humans achieve task mastery on VisuLogic with hints. Value changes are color-coded with \textcolor{red}{red} indicating positive shifts and \textcolor{reduce-color}{green} denoting negative variations.}
  \tiny
  \label{hint_table}
  \centering
    \resizebox{\linewidth}{!}{
  \begin{tabular}{lllllllll}
    \toprule
    \textbf{Models} & \textbf{Hint} & \textbf{Overall} & \textbf{Quantity} & \textbf{Spatiality} & \textbf{Position} & \textbf{Attribute} & \textbf{Style} & \textbf{Other}  \\
    \midrule
    \midrule
    \multirow{2}{*}{Human} &\XSolid  & 51.4 & 45.3 & 52.7 & 71.1 & 50.0 & 47.5 & 44.2 \\
    &\Checkmark & $83.6_{\textcolor{red}{(+32.2)}}$ & $85.1_{\textcolor{red}{(+39.8)}}$ & $68.5_{\textcolor{red}{(+15.8)}}$ & $100.0_{\textcolor{red}{(+28.9)}}$ & $95.7_{\textcolor{red}{(+45.7)}}$ & $78.6_{\textcolor{red}{(+31.1)}}$ & $90.5_{\textcolor{red}{(+46.3)}}$ \\
     \midrule
   \multirow{2}{*}{GPT-4o (20240806)} &\XSolid  & 26.3 & 28.6 & 24.7 & 27.2 & 26.8 & 20.0 & 25.9 \\
    &\Checkmark & $30.0_{\textcolor{red}{(+3.7)}}$ & $25.4_{\textcolor{reduce-color}{(-3.2)}}$ & $31.5_{\textcolor{red}{(+6.8)}}$ & $29.2_{\textcolor{red}{(+2.0)}}$ & $28.6_{\textcolor{red}{(+1.8)}}$ & $30.8_{\textcolor{red}{(+10.8)}}$ & $42.9_{\textcolor{red}{(+17.0)}}$ \\
     \midrule
   \multirow{2}{*}{Claude-3.7-Sonnet (20250219)} &\XSolid & 24.8 & 22.7 & 27.3 & 27.9 & 28.0 & 22.2 & 22.2 \\
    &\Checkmark & $33.5_{\textcolor{red}{(+8.7)}}$ & $37.3_{\textcolor{red}{(+14.6)}}$ & $33.3_{\textcolor{red}{(+6.0)}}$ & $37.5_{\textcolor{red}{(+9.6)}}$ & $23.8_{\textcolor{reduce-color}{(-4.2)}}$ & $15.4_{\textcolor{reduce-color}{(-6.8)}}$ & $38.1_{\textcolor{red}{(+15.9)}}$ \\
     \midrule
   \multirow{2}{*}{Gemini-2.0-Pro (20250205)} &\XSolid & 28.0 & 29.7 & 24.2 & 27.9 & 30.5 & 22.2 & 33.3 \\
    &\Checkmark  & $36.5_{\textcolor{red}{(+8.5)}}$ & $44.8_{\textcolor{red}{(+15.1)}}$ & $33.3_{\textcolor{red}{(+9.1)}}$ & $25.0_{\textcolor{reduce-color}{(-2.9)}}$ & $38.1_{\textcolor{red}{(+7.6)}}$ & $15.4_{\textcolor{reduce-color}{(-6.8)}}$ & $42.9_{\textcolor{red}{(+9.6)}}$ \\
     \midrule
   \multirow{2}{*}{Doubao-1.5-Vision-Pro-32k (20250115)} &\XSolid & 28.1 & 28.1 & 23.8 & 29.1 & 25.1 & 32.1 & 35.0 \\
    &\Checkmark & $37.0_{\textcolor{red}{(+8.9)}}$ & $46.3_{\textcolor{red}{(+18.2)}}$ & $25.9_{\textcolor{red}{(+2.1)}}$ & $54.2_{\textcolor{red}{(+25.1)}}$ & $33.3_{\textcolor{red}{(+8.2)}}$ & $23.1_{\textcolor{reduce-color}{(-9.0)}}$ & $28.6_{\textcolor{reduce-color}{(-6.4)}}$ \\
    \bottomrule
  \end{tabular}}
  \vspace{-0.5em}
\end{table}

\begin{figure}[h]
    \centering
    \includegraphics[width=1\linewidth]{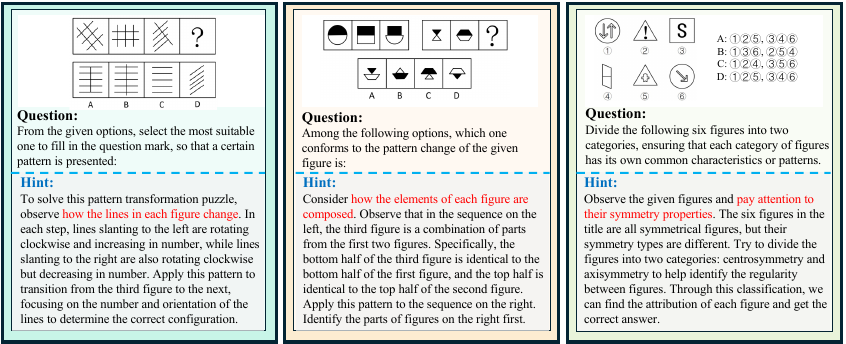}
    \caption{\textbf{Hint prompts visualization.} Hint prompts examples, which supply solution guidance for MLLMs, are shown in the image, with solution-critical elements highlighted in \textcolor{red}{red}.}
    \label{fig:hint_example}
    \vspace{-0.5em}
\end{figure}

\section{Experiments}

In this section, we present a comprehensive evaluation of the VisuLogic benchmark. We first describe the experimental setup in Section~\ref{sec:ExperimentSetup}, followed by overall performance results in Section~\ref{sec:overall-results}. We then analyze systematic errors in Section~\ref{sec:error-analysis} and provide qualitative insights in Section~\ref{sec:qualitative-analysis}.

\subsection{Experiment Setup}
\label{sec:ExperimentSetup}

\noindent \textbf{References Performance.}
To fully investigate models' performance, we establish two reference points: 1) \textbf{Human Performance}: We invited 100 graduate students majoring in science and engineering to solve 10 randomly sampled VisuLogic questions each, allowing 2–5 minutes per question. The aggregate accuracy over all participants constitutes the human benchmark.  2) \textbf{Random Selection}: We simulate random guessing by sampling answers uniformly over 10 independent runs and report the average accuracy as the random baseline.

\vspace{0.5em}
\noindent \textbf{Evaluated Models.} We evaluate a total of 28  models on VisuLogic, comprising 8 large language models (LLMs) and 20 multimodal large language models (MLLMs).  For open-source LLMs, we test \textit{Deepseek-R1}~\cite{deepseekai2025deepseekr1incentivizingreasoningcapability}, \textit{Qwen2.5-72B-Instruct}~\cite{qwen2.5} and \textit{Qwen-QwQ}~\cite{qwq32b}, and for close-source LLMs we evaluate \textit{GPT-4}~\cite{achiam2023gpt}, \textit{o3-mini}, \textit{Gemini-2.0-Flash-Thinking}~\cite{team2023gemini}, \textit{Claude-3.7-Sonnet} and \textit{Doubao-1.5-Pro-32k}. Open-source MLLMs include \textit{Qwen2.5-VL-7B-Instruct }~\cite{Qwen2.5-VL}, \textit{Qwen2.5-VL-72B-Instruct}~\cite{Qwen2.5-VL}, \textit{QvQ-72B-Preview}~\cite{qvq-72b-preview}, \textit{InternVL2.5-38B}~\cite{chen2024internvl}, \textit{InternVL2.5-78B}~\cite{chen2024internvl}, \textit{InternVL3-38B}~\cite{zhu2025internvl3exploringadvancedtraining},  \textit{InternVL3-78B}~\cite{zhu2025internvl3exploringadvancedtraining}, \textit{LLaVA-v1.5-7B}~\cite{liu2023llava}, \textit{LLaVA-OneVision-7B (SI)}~\cite{li2024llavaonevisioneasyvisualtask}, \textit{ShareGPT4V}~\cite{chen2023sharegpt4v}, \textit{MiniCPM-o-2.6}~\cite{yao2024minicpm}, \textit{GLM-4v-9B}~\cite{glm2024chatglm}, \textit{Ovis2-8B}~\cite{lu2024ovis} and \textit{mPLUG-Owl3-7B}~\cite{ye2024mplugowl3longimagesequenceunderstanding}, while close-source MLLMs consist of \textit{GPT-4o}~\cite{hurst2024gpt}, \textit{GPT-4o-mini}, \textit{Kimi-latest}~\cite{team2025kimi}, \textit{Doubao-1.5-Vision-Pro-32k}, \textit{Gemini-2.0-Pro}~\cite{team2023gemini} and \textit{Claude-3.7-Sonnet}. 
We further include two reinforcement-learning baselines built on \textit{Qwen2.5-VL-7B-Instruct}~\cite{Qwen2.5-VL} and \textit{InternVL2.5-38B}~\cite{chen2024internvl}, respectively, trained via our rule-based RL procedure on our supplementary training dataset. Fully supervised fine-tuning (SFT) experiments on the same datasets serve as controls to isolate the effect of RL optimization.
All model hyperparameters, training regimes, and implementation details are provided in the Appendix.

\vspace{0.5em}

\noindent\textbf{LLM Evaluation Protocol.} For language-only models, we generate an auxiliary image description using GPT-4o and prepend it to the question. Specifically, each question is formatted as ``\textit{Following is a detailed caption describing an image:  [IMAGE DESCRIPTION]. Based on the provided description, select the best answer from the four options:}''.
This combined prompt is fed directly into the target LLMs for inference.

\noindent\textbf{Prompts Setting.} 
We apply three distinct prompting paradigms to investigate model reasoning capabilities: 
1)~\textbf{Non-CoT prompt evaluation}: Models receive a concise instruction:  \enquote{\textit{Answer the question using a single word or phrase, following this format: Answer: \textbackslash boxed\{\$LETTER\}}}.
2)~\textbf{CoT prompt evaluation}: We prompt models to articulate intermediate reasoning steps: \enquote{\textit{Solve the complex visual logical reasoning problem through step-by-step reasoning. Think about the reasoning process first and answer the question following this format: Answer: \textbackslash boxed\{\$LETTER\}}}.
3)~\textbf{Hint prompts evaluation}: Leveraging GPT-4o, we generate question-specific hints derived from the reference solutions. As shown in Figure~\ref{fig:hint_example}, solution-related hints are provided alongside the CoT prompt to guide reasoning without revealing the final answer directly. 
Notably, unless otherwise specified, CoT prompt evaluation is employed by default for assessing model performance.

\subsection{Overall Results} \label{sec:overall-results}

\noindent \textbf{LLM Performance.}
Table~\ref{sample-table-8x8} reports that all evaluated LLMs attain rather low accuracy on VisuLogic. The best-performing LLM, \textit{Qwen2.5-72B-Instruct}, reaches only 28.0\%, while \textit{GPT-4} and \textit{Deepseek-R1} achieve 23.6\% and 26.6\%, respectively. These findings underscore that reasoning based solely on textual descriptions is insufficient to capture the rich visual information required by our benchmark, causing failures to resolve  visual logical reasoning problems.

\vspace{0.5em}

\noindent \textbf{MLLM Performance.} As shown in Table~\ref{sample-table-8x8}, current multimodal LLMs also perform poorly on VisuLogic. The highest score is 28.1\% by \textit{Doubao-1.5-Vision-Pro-32k}, which remains a substantial 23.3 points below human performance. Advanced models such as \textit{GPT-4o} and \textit{Gemini-2.0-Pro} attain only 26.3\% and 28.0\%, respectively, revealing a marked gap between existing MLLMs and human-level visual reasoning. Overall, these results indicate that current MLLMs have serious deficiencies in visual reasoning and that significant advances are still required.

\vspace{0.5em}

\noindent \textbf{Effectiveness of CoT Prompts.}
Contrary to expectations, chain-of-thought (CoT) prompting yields minimal improvements in visual reasoning. As detailed in Table~\ref{cot_table}, \textit{GPT-4o-mini} benefits most, with only a 1.2-point gain under CoT compared to direct-answer prompts; all other models exhibit gains below 1.0 point. We speculate that this limited effect likely stems from current CoT training being based only on pure-text corpora; future works should explore CoT techniques tailored to multimodal data to better support visual reasoning tasks.

\vspace{0.5em}

\noindent \textbf{Effectiveness of Hint Prompts.} 
Table~\ref{hint_table} shows that hint prompts can boost model performance—\textit{Claude-3.7-Sonnet}, \textit{Gemini-2.0-Pro}, and \textit{Doubao-1.5-Vision-Pro-32k} all improve by over 8 points, reaching accuracies above 35\%. However, even with explicit guidance, models still fail to construct coherent, reliable reasoning chains. This suggests that simply augmenting training data with similar tasks is insufficient (which can help MLLMs come up with specific directions for solving the problem); future efforts must focus on enhancing the reliability and correctness of reasoning procedures of MLLMs to achieve more accurate reasoning inference.

\vspace{0.5em}
\noindent \textbf{Impact of Model Scaling.} 
In Table~\ref{sample-table-8x8}, we observe a positive correlation between parameter size and model performance. With in the same model series, \textit{Qwen2.5-VL-72B-Instruct} achieves 26.2 \% outperforming \textit{Qwen2.5VL-7B-Instruct} (26.0\%) by 0.2\%. Furthermore, \textit{InternVL2.5-78B} (27.3\%) surpasses \textit{InternVL2.5-38B} (25.5\%) by a margin of 1.8\%.

\vspace{0.5em}

\noindent \textbf{Open-Source vs Close-Source.} 
Table~\ref{sample-table-8x8} further compares open- and closed-source models. The top open-source MLLM, \textit{InternVL3-78B}, attains 27.7\%, trailing the closed-source leader (\textit{Doubao-1.5-Vision-Pro-32k}, 28.1\%) by only 0.4\% points and outperforming other proprietary competitors such as \textit{GPT-4o} and \textit{Claude-3.7-Sonnet}. Overall, both open- and closed-source models exhibit uniformly low performance, highlighting a widespread neglect of visual reasoning objectives in current multimodal model training and data collection.

\vspace{0.5em}

\noindent \textbf{Behaviors of RL Trained models.} 
As shown in Table~\ref{sample-table-8x8}, MLLMs with reinforcement learning optimization can yield obvious improvements in visual reasoning performance. \textit{Qwen2.5-VL-7B-Instruct-RL} attains 28.0\%, a 2.0 percentage point boost over its non-RL counterpart. More strikingly, \textit{InternVL2.5-38B-RL} reaches 31.1\%, surpassing the original non-RL model by 5.6\% and establishing a new state-of-the-art on VisuLogic. Furthermore, compared to supervised fine-tuning (SFT) on identical datasets, RL-enhanced models demonstrate substantially larger performance gains, underscoring the promise of targeted RL methods for advancing multimodal visual reasoning.

\subsection{Fine-grained Comparison} \label{sec:error-analysis}
\begin{figure}[tbp]  
    \hspace{-0.5cm}
    \centering
    \begin{subfigure}{0.325\linewidth}
		\centering
		\includegraphics[width=1.05\linewidth]{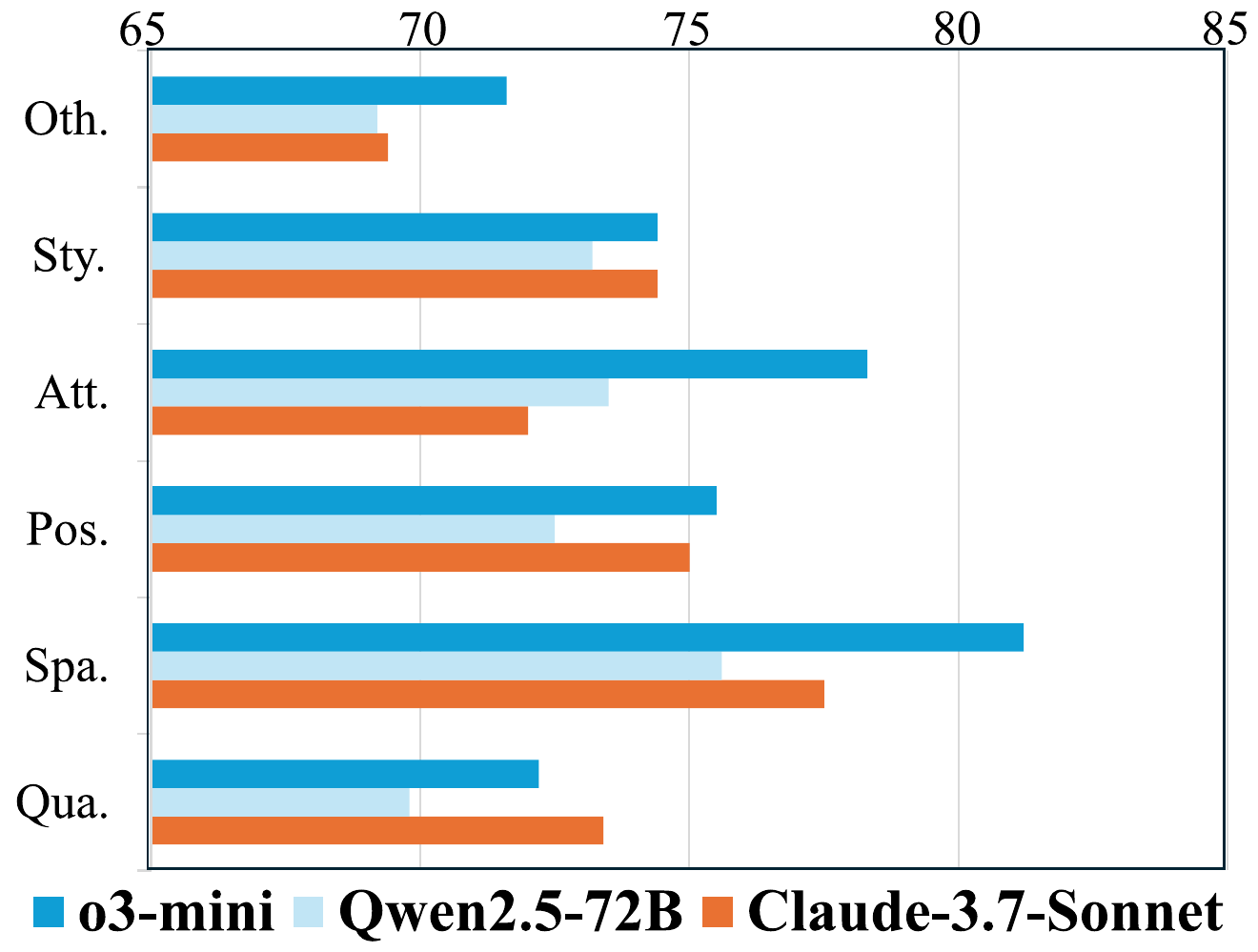}
		\caption{LLMs' error distribution.}
		\label{fig:error_distribute_3-1}
	\end{subfigure}
    \begin{subfigure}{0.329\linewidth}
		\centering
		\includegraphics[width=1.05\linewidth]{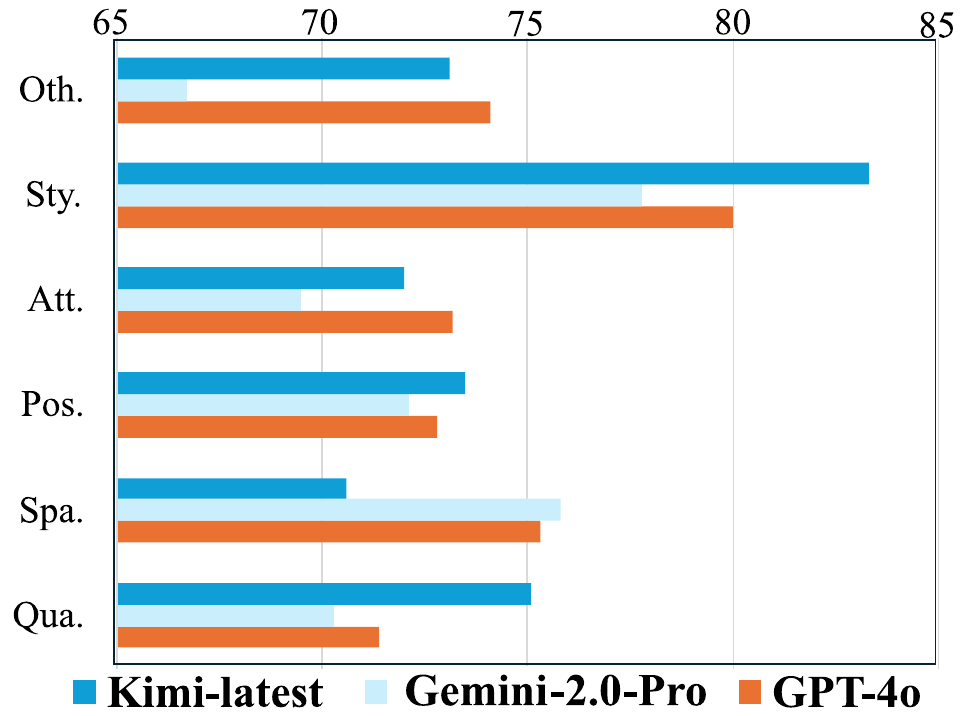}
		\caption{MLLMs' error distribution.}
		\label{fig:error_distribute_3-2}
	\end{subfigure}
    \begin{subfigure}{0.327\linewidth}
		\centering
		\includegraphics[width=1.06\linewidth]{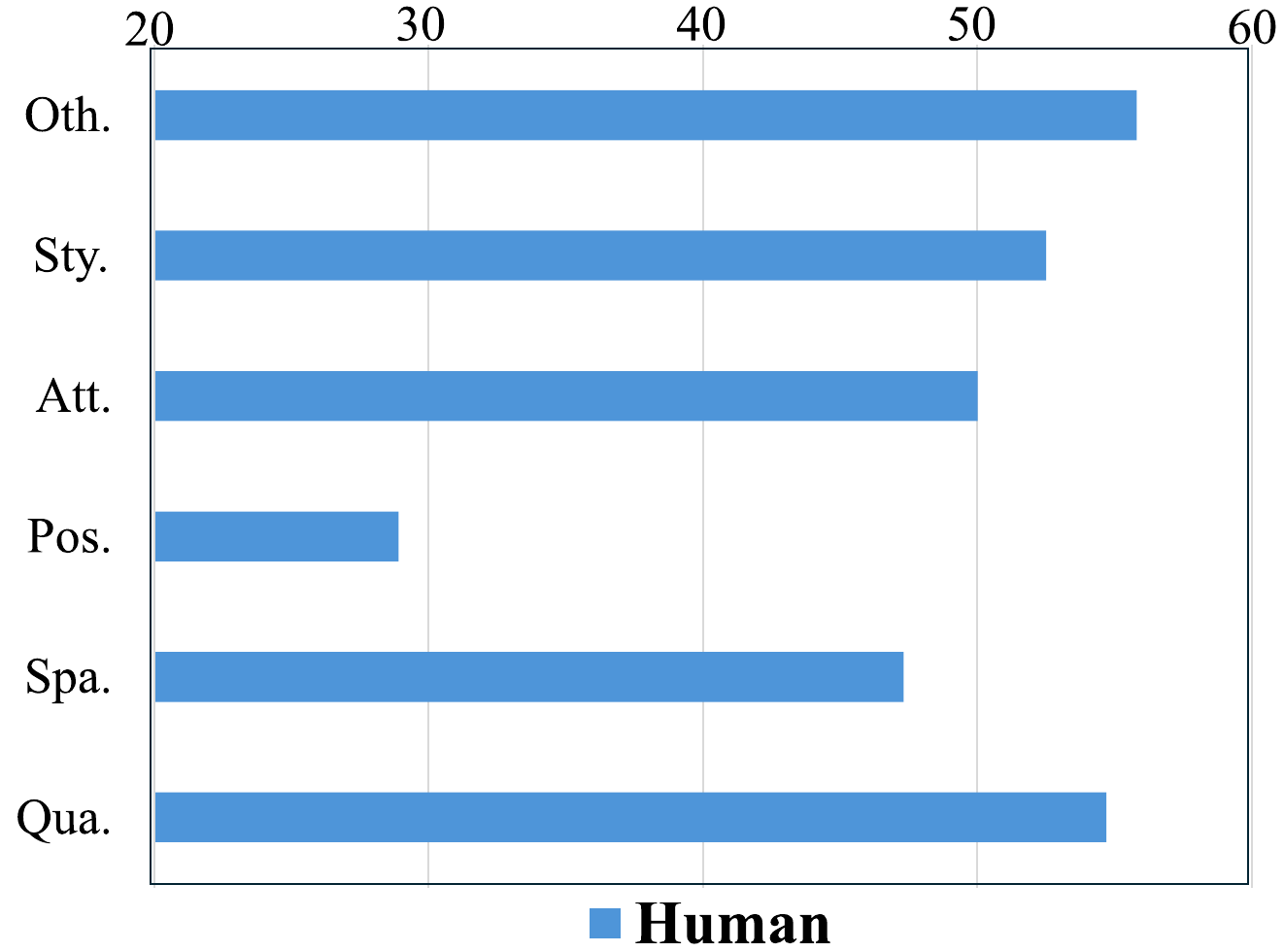}
		\caption{Humans' error distribution.}
		\label{fig:error_distribute_3-3}
	\end{subfigure}
    \caption{\textbf{Error distribution analysis.} The figure demonstrates distinct error type allocations across Humans, LLMs and MLLMs, revealing differences among their cognition patterns.}  
    \label{fig:distribution}  
    \vspace{-0.5em}
\end{figure}

We systematically analyze model capabilities by examining error distributions across reasoning categories for different models. Figure~\ref{fig:distribution} presents the error rates of LLMs, MLLMs, and human participants over six distinct reasoning categories.

Figure~\ref{fig:error_distribute_3-1} reveals that LLMs struggle most with \textit{Spatial Reasoning} questions, indicating that text-only descriptions are insufficient to infer three-dimensional structures or spatial transformations. In contrast, their performance on \textit{Quantitative Reasoning} tasks is comparatively stronger, suggesting that quantitative relationships are more readily conveyed through language.

As shown in Figure~\ref{fig:error_distribute_3-2}, \textit{Stylistic Reasoning} presents the greatest difficulty for MLLMs, with error rates exceeding 75\%—worse than random guessing (25\% accuracy). This result underscores a fundamental limitation of current MLLM architectures in capturing subtle visual cues such as overlays, contours, and shape variations.

Figure~\ref{fig:error_distribute_3-3} reveals that human error patterns form a distinct cluster, separate from both LLMs and MLLMs. Human participants maintain error rates below 30\% on \textit{Positional Reasoning} tasks, reflecting robust position-based visual inference. By contrast, both model classes struggle with positional reasoning, highlighting a fundamental divergence in visual–cognitive processes between humans and MLLMs.

\subsection{Qualitative Analysis} \label{sec:qualitative-analysis}

\noindent \textbf{LLM Failures.} As shown in Figure~\ref{fig:5_model_output}(h), text-only LLMs that rely on externally generated image captions often omit critical visual details required for multi-step logical deduction—such as the counts, shapes, and progression patterns of the black and white dots in Figure~\ref{fig:5_model_output}(a). Consequently, their reasoning diverges from the correct solution and frequently yields hallucinations or irrelevant responses.

\noindent \textbf{MLLM Failures.} Figure~\ref{fig:5_model_output} also presents cases in which MLLMs correctly describe static visual content yet fail to infer the evolving relationships among shapes, instead resorting to superficial cues like object counts. While these models can recognize individual shapes and tally items, they struggle to reason over inter-element relations, which limits their ability to solve complex visual-logic problems.

\noindent \textbf{RL-Based Improvements.} As illustrated in Figure~\ref{fig:5_model_output}(g), reinforcement learning (RL) encourages deeper, stepwise logical reasoning. The RL-enhanced model successfully captures state transitions (e.g., the movements of chess pieces in Figure~\ref{fig:5_model_output}(a)) and accurately predicts subsequent configurations. Moreover, it learns to iteratively revise intermediate hypotheses—akin to trial-and-error—until a coherent deduction emerges (see additional examples in the Appendix). These findings highlight the potential of RL methods to bolster performance on visual reasoning tasks.

\section{Conclusion}
In this paper, we present VisuLogic, a novel benchmark designed to evaluate the visual reasoning capabilities of Multi-modal Large Language Models (MLLMs). The benchmark consists of 1,000 vision-centric reasoning tasks distributed across six distinct categories. We conduct  comprehensive evaluation of several advanced LLMs and MLLMs on this benchmark and provide an in-depth analysis of their performance. Our findings reveal that even the most advanced models fall short of human performance, highlighting substantial opportunities for advancement in visual logical reasoning. Through  further experiments, we find that reinforcement learning (RL) is a promising approach for enhancing the vision reasoning capabilities of MLLMs. To promote further research and innovation, we open-source the evaluation code, training scripts, and datasets associated with this work.

\newpage
\bibliographystyle{abbrv}
\bibliography{ref}
\newpage
\appendix
\section{Overview of the Appendix}
In the appendix, we provide additional details and supplementary information to further elaborate on sections mentioned above.
In Section~\ref{apdx:benchmark}, we analyze the statistical features of the dataset, meanwhile providing examples of questions ranging from different categories.
Section~\ref{apdx:evaluation} contains experiments details, including the evaluation of LLMs, the evaluation of hint prompts and RL experiments. Some examples of model outputs are also illustrated.


\section{Benchmark Analysis}\label{apdx:benchmark}

\subsection{Statistical analysis}
\begin{figure}[h]
    \centering
    \includegraphics[width=0.5\linewidth]{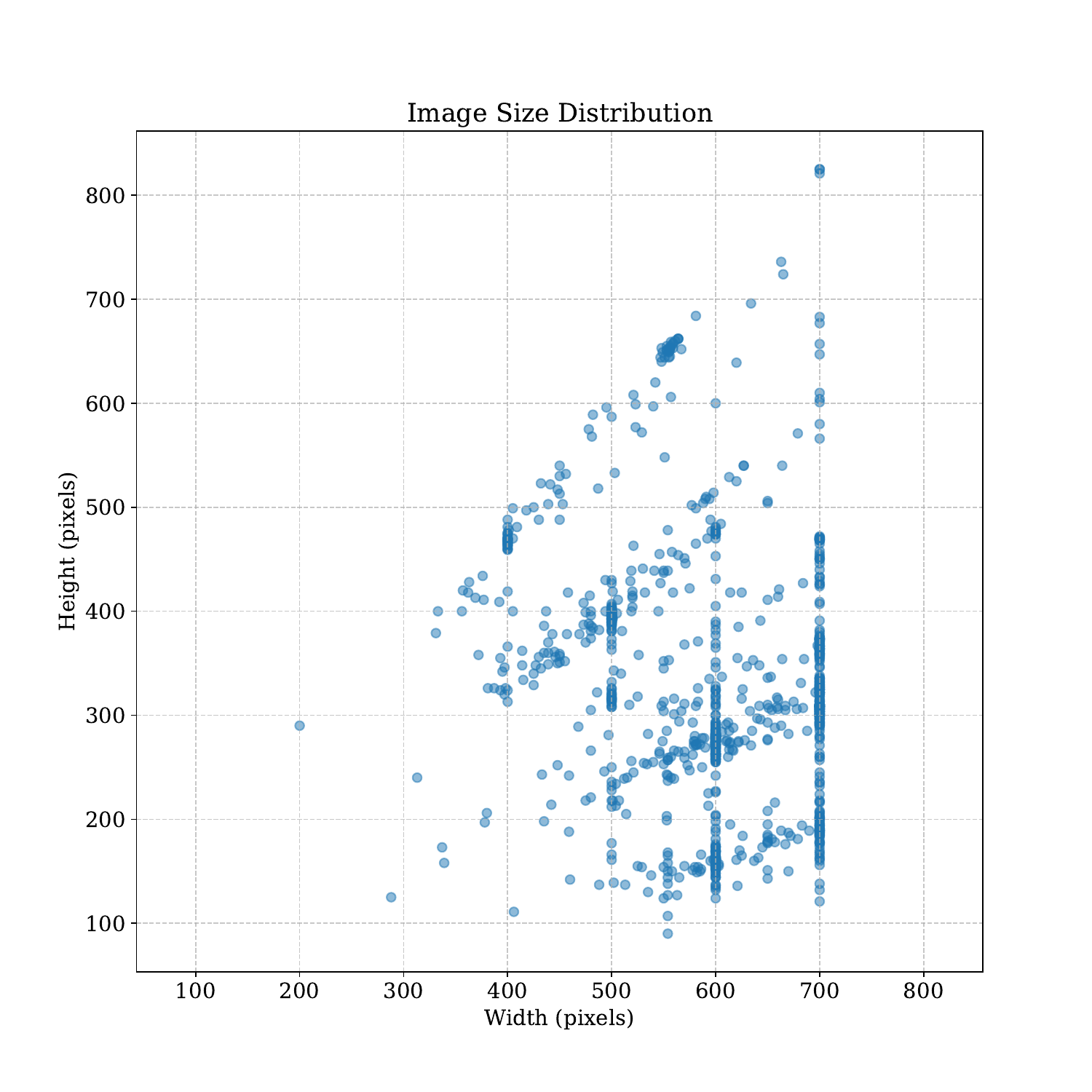}
    \caption{\textbf{Image size distribution.} The size of images is limited to within the same order of magnitude.}
    \label{fig:appendix_image_size_distribution}
\end{figure}

\begin{figure}[ht]
    \centering
    \includegraphics[width=1\linewidth]{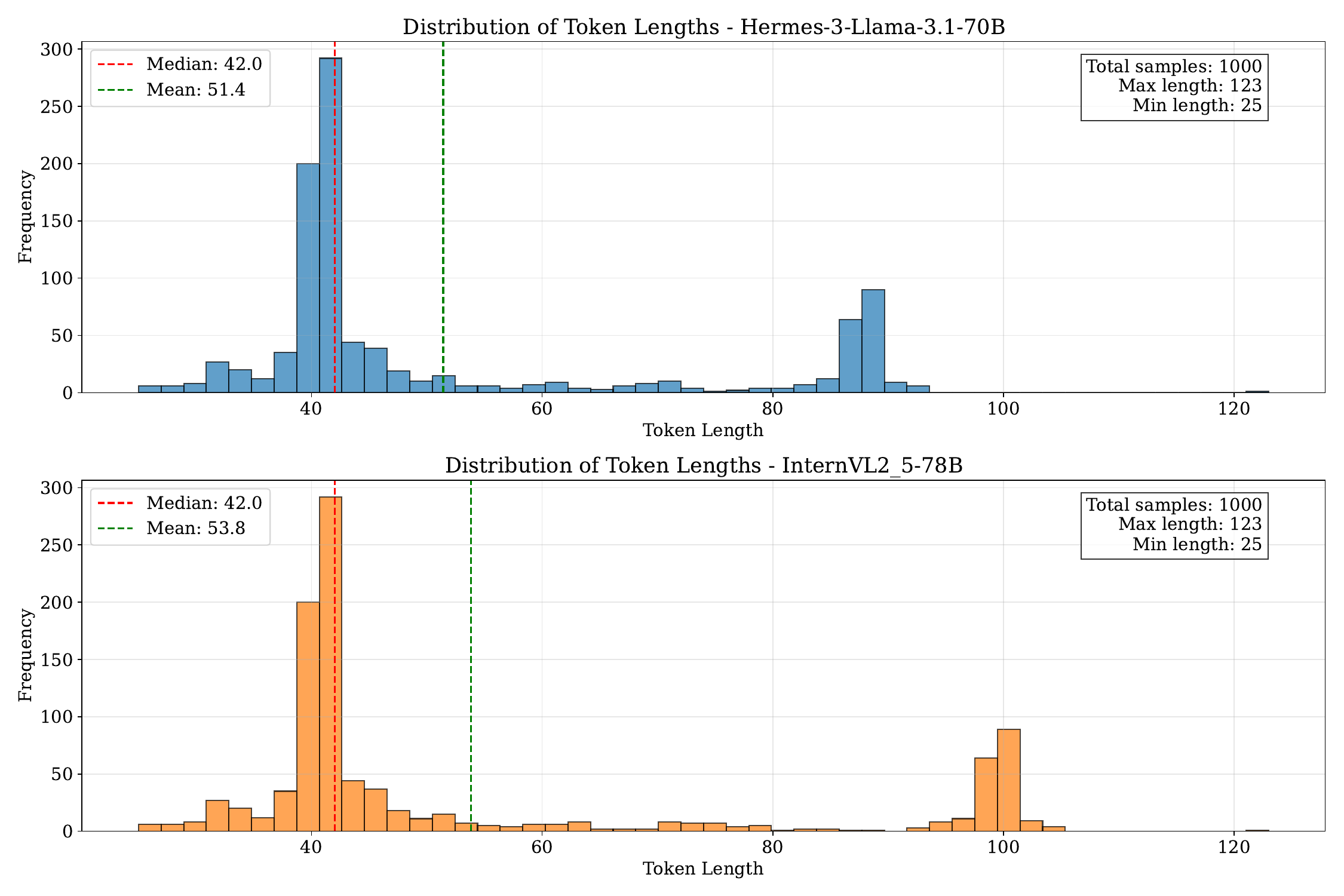}
    \caption{Distribution of text token length in VisuLogic.}
    \label{fig:appendix_text_len_distribution}
\end{figure}

As shown in Figure~\ref{fig:appendix_text_len_distribution}, the text length of questions in VisuLogic is mostly concentrated around 40 tokens (calculated by Llama-3.1's and  InternVL2.5's tokenizer). We also analyze the distribution of image sizes, as shown in Figure~\ref{fig:appendix_image_size_distribution}. The image widths range from 200 to 700 pixels, with an average of 592.3 pixels, while the heights range from 90 to 825 pixels, with an average of 327.9 pixels.
\subsection{More Examples of VisuLogic}

\begin{figure}
    \centering
    \includegraphics[width=1.05\linewidth]{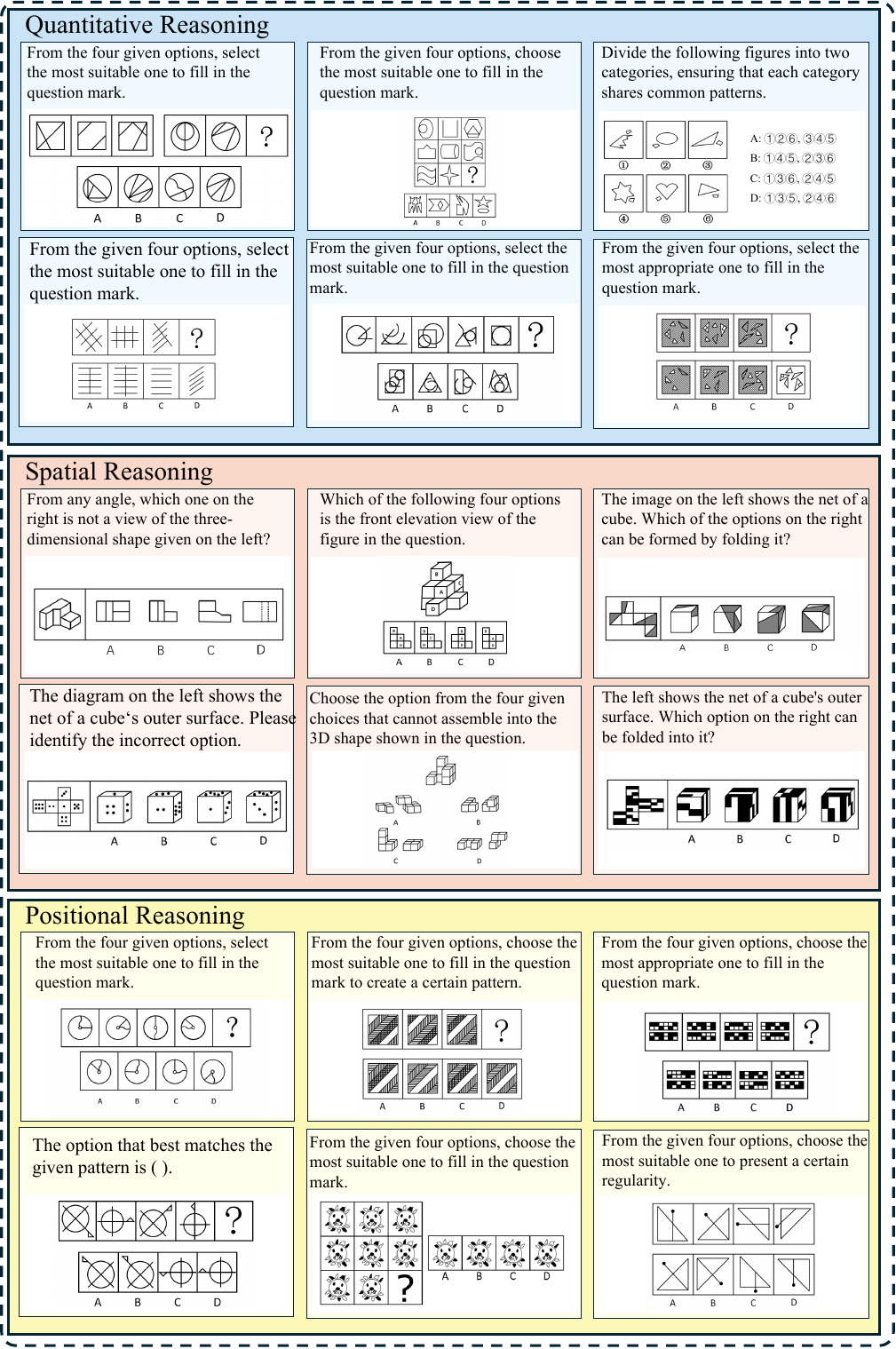}
    \caption{More examples in \benchmarkshortname of Quantitative Reasoning, Spatial Reasoning, Positional Reasoning.}
    \label{fig:examples_in_benchmarks_1}
\end{figure}

\begin{figure}
    \centering
    \includegraphics[width=1.05\linewidth]{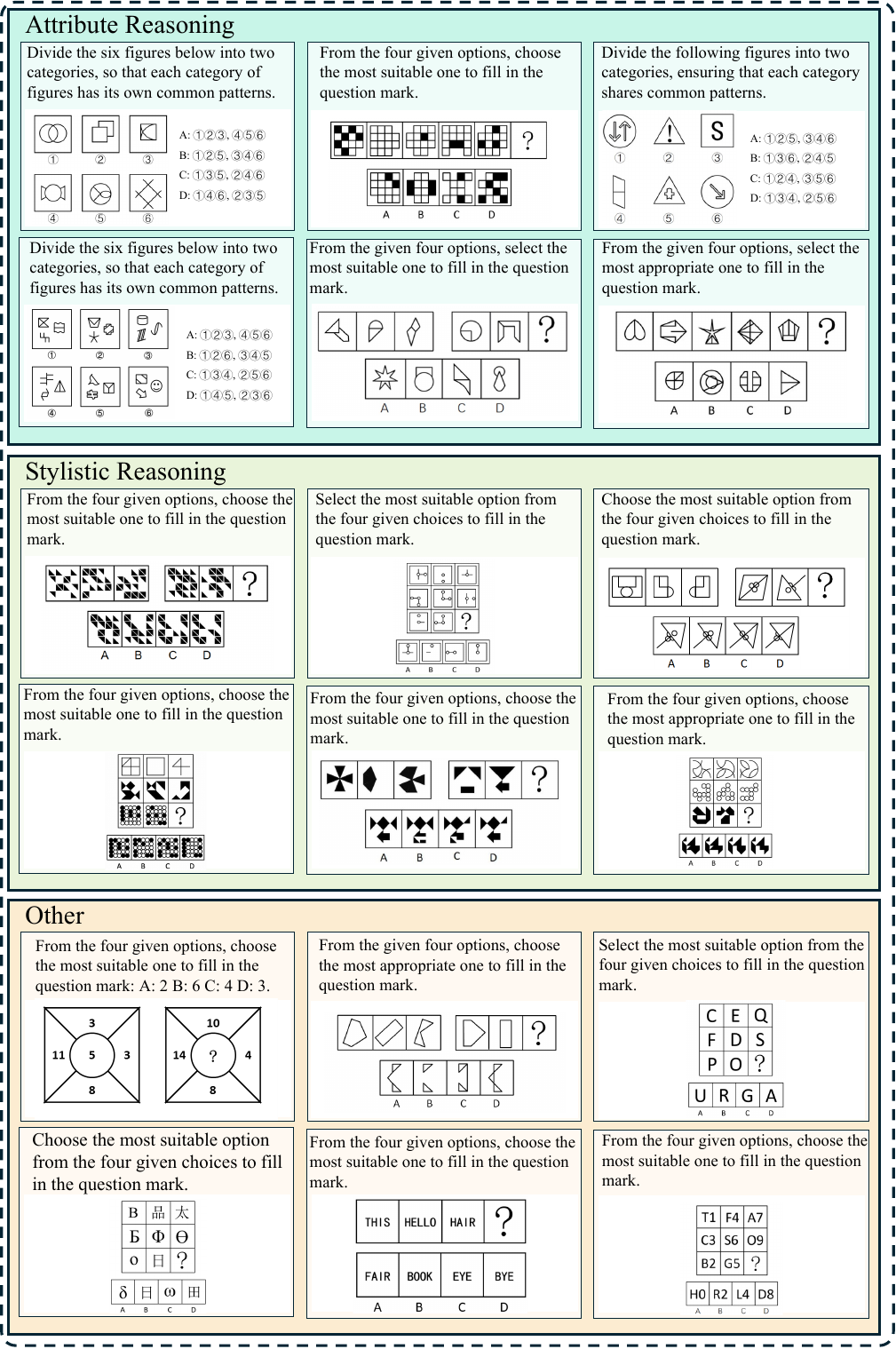}
    \caption{More examples in \benchmarkshortname of Attribute Reasoning, Stylistic Reasoning, and Other.}
    \label{fig:examples_in_benchmarks_2}
\end{figure}

To provide a thoroughly presentation of our benchmark, we include more examples of questions from different categories in the Figure~\ref{fig:examples_in_benchmarks_1} and Figure~\ref{fig:examples_in_benchmarks_2}. 


\section{Evaluation \& Experiment}\label{apdx:evaluation}
\subsection{Evaluation of LLMs}
\begin{figure}[h]
    \centering
    \includegraphics[width=1\linewidth]{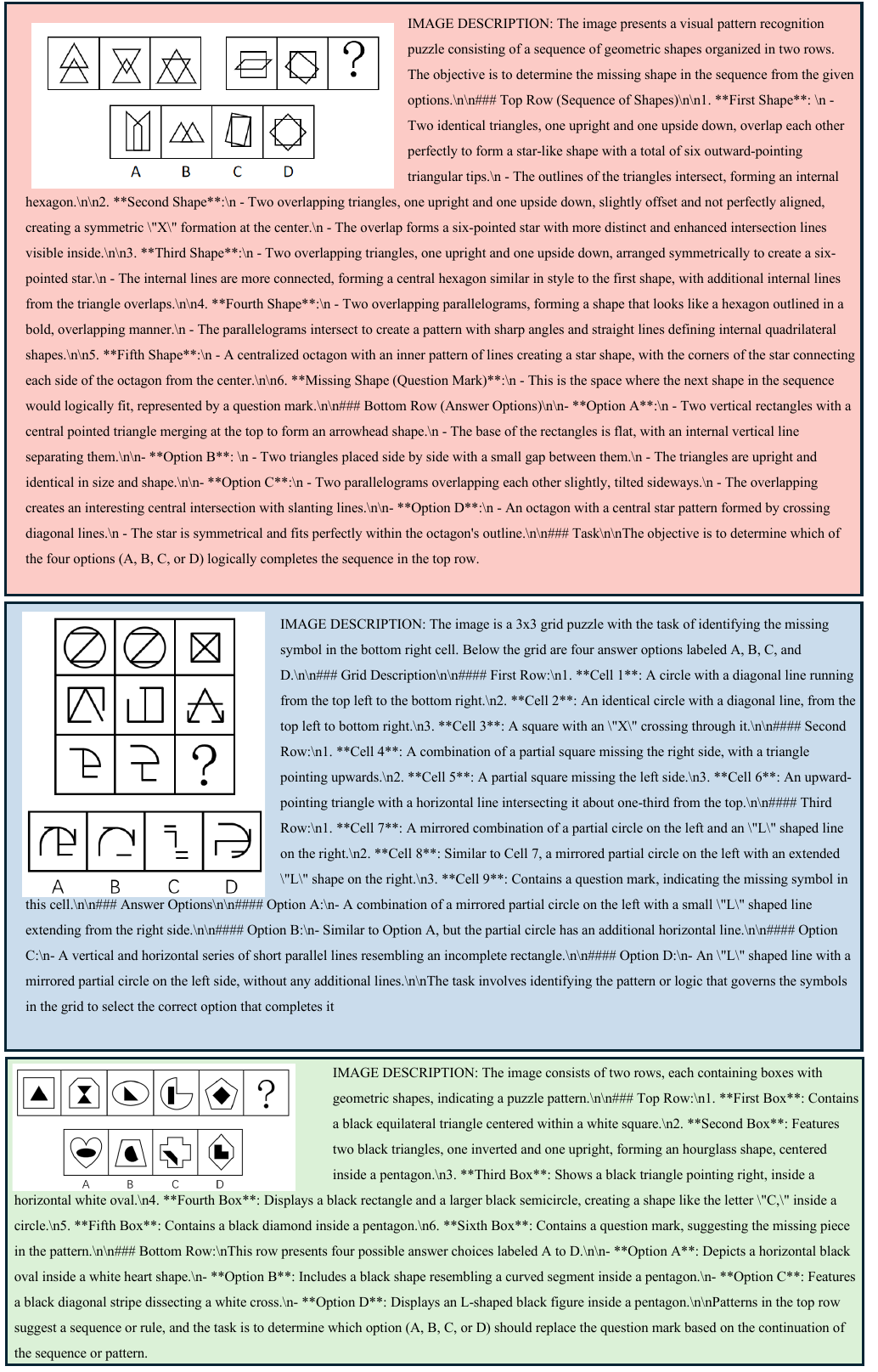}
    \caption{Part of image caption in LLM evaluation.}
    \label{fig:image_cation1}
\end{figure}

\begin{figure}[h]
    \centering
    \includegraphics[width=1\linewidth]{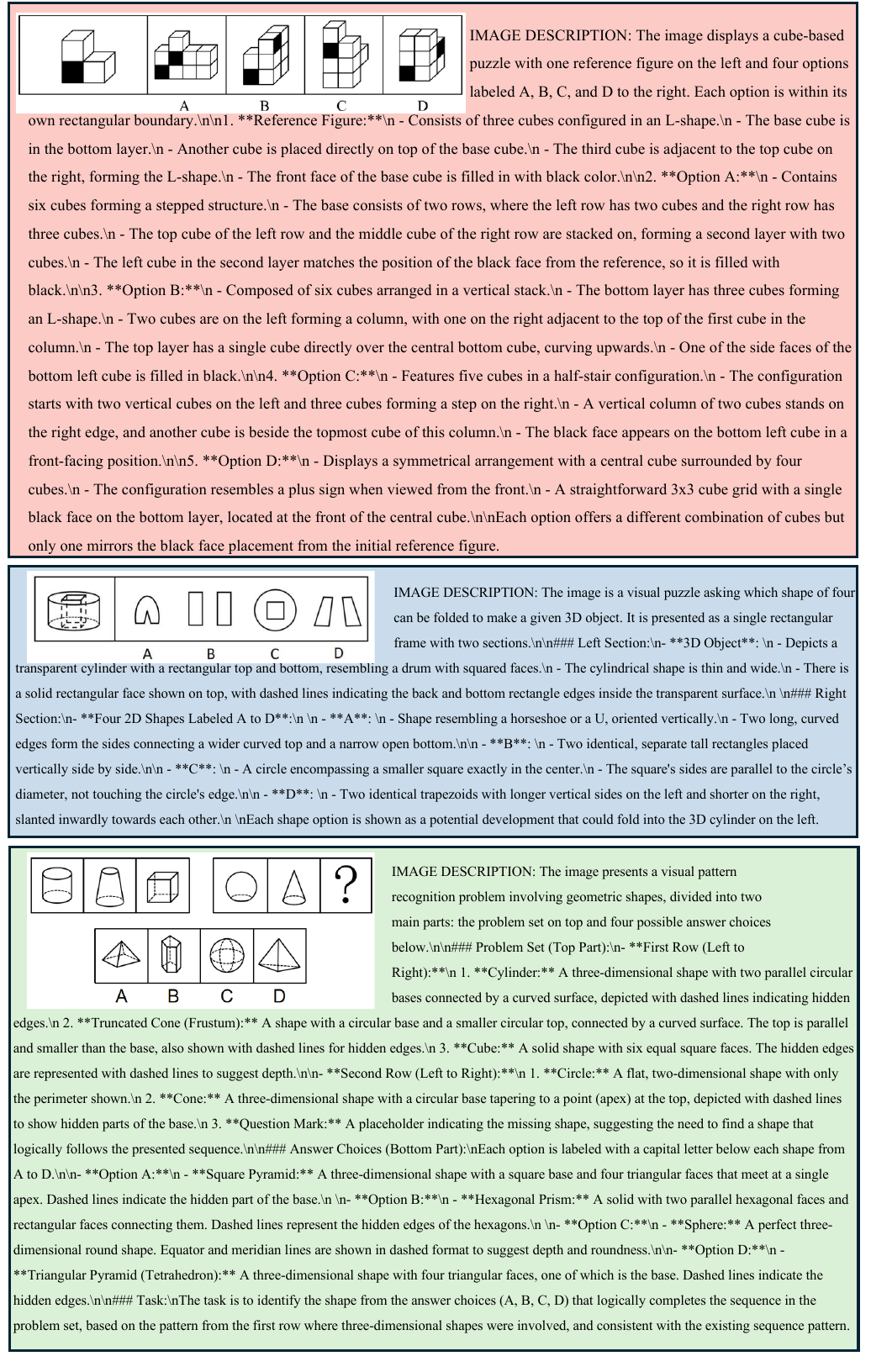}
    \caption{Part of image caption in LLM evaluation.}
    \label{fig:image_cation2}
\end{figure}

\noindent\textbf{Caption generation for LLMs Evaluation.}
In our experiment, we employ large language models (LLMs) for comparative analysis. Specifically, when setting up the LLM-based experiment, we initially utilize GPT-4o to generate captions for images with the following prompt: \textit{Please describe the fine-grained content of the image or figure based on this question, including scenes, objects, relationships, and any text present. Please note that you do not need to answer this question directly, just describe the information of this picture}. Additional examples of generated image captions are presented in Figure~\ref{fig:image_cation1} and Figure~\ref{fig:image_cation2}.

\noindent\textbf{More Examples of Captions.}
We provide additional image captions for six categories, as illustrated in Figures~\ref{fig:image_cation1} and \ref{fig:image_cation2}. Even SOTA MLLM (GPT-4o) encounters difficulties in accurately describing the details of images from  VisuLogic. 

\begin{figure}[h]
    \centering
    \includegraphics[width=1\linewidth]{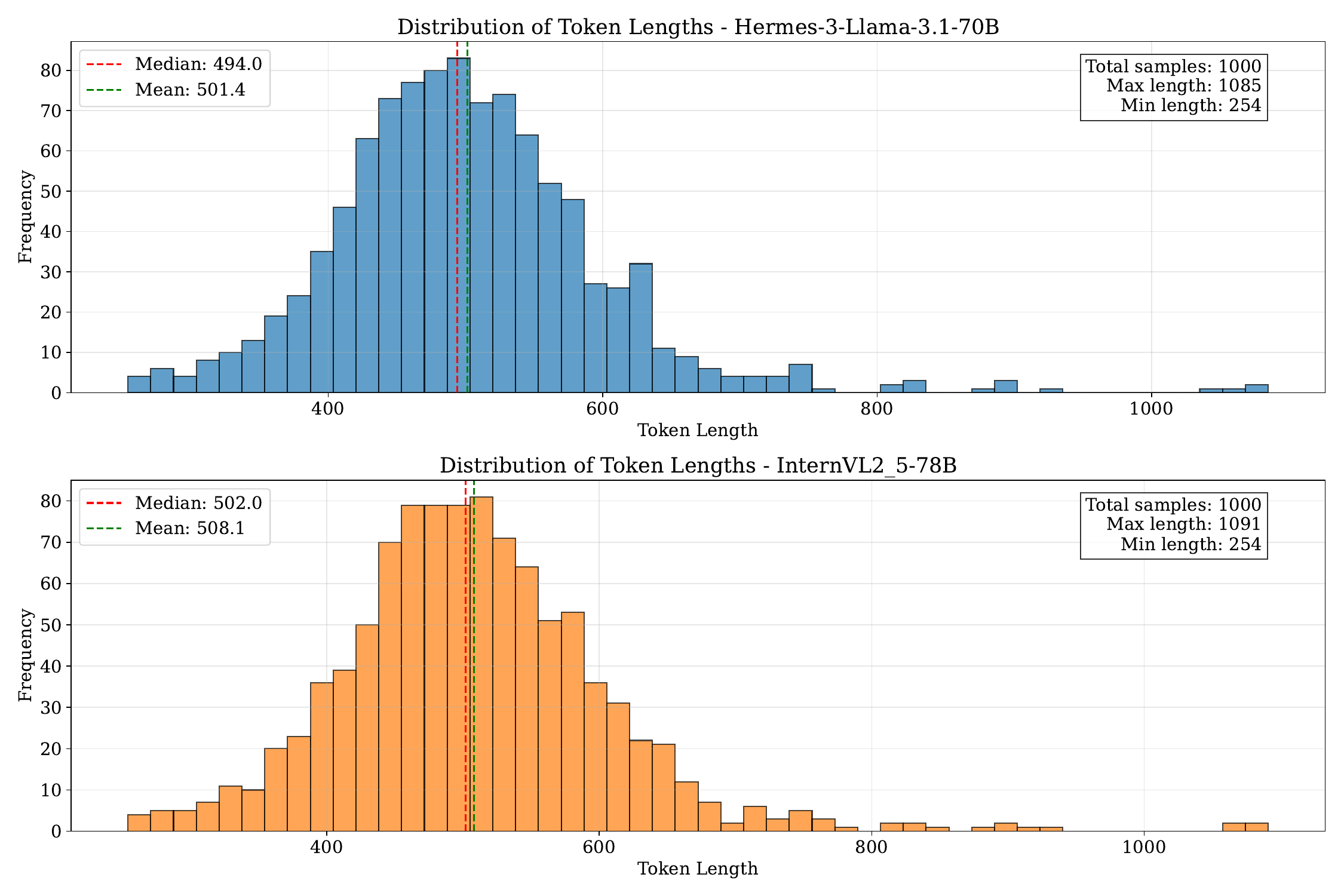}
    \caption{Distribution of tokens length in LLM evaluation settings, including image description.}
    \label{fig:benchmark_des_len_comparison}
\end{figure}


\subsection{More Solutions from Models}
\begin{figure}
    \centering
    \includegraphics[width=1\linewidth]{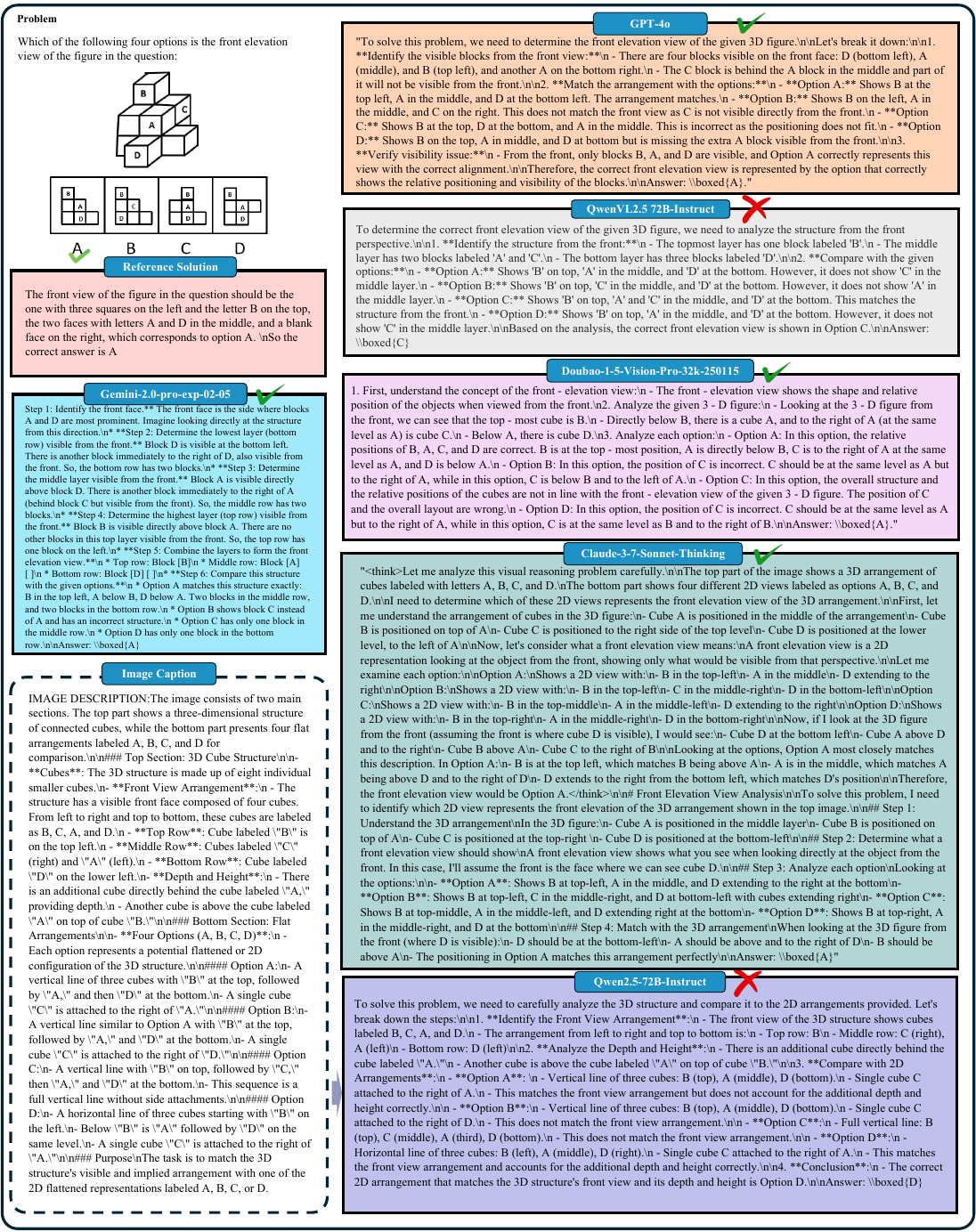}
    \caption{\textbf{Solution examples generated by different models.} Reference solution and outputs generated by GPT-4o, Qwen2.5VL-72B-Instruct, Gemini-2.0-pro-exp-02-05, Doubao-1.5-Vision-Pro-32K and Claude-3.7-sonnet-thinking. Additionally, the image caption and solution from LLMs (Qwen2.5-72B-Instruct) are also illustrated.}
    \label{fig:appendix_model_output_1}
\end{figure}

\begin{figure}
    \centering
    \includegraphics[width=1\linewidth]{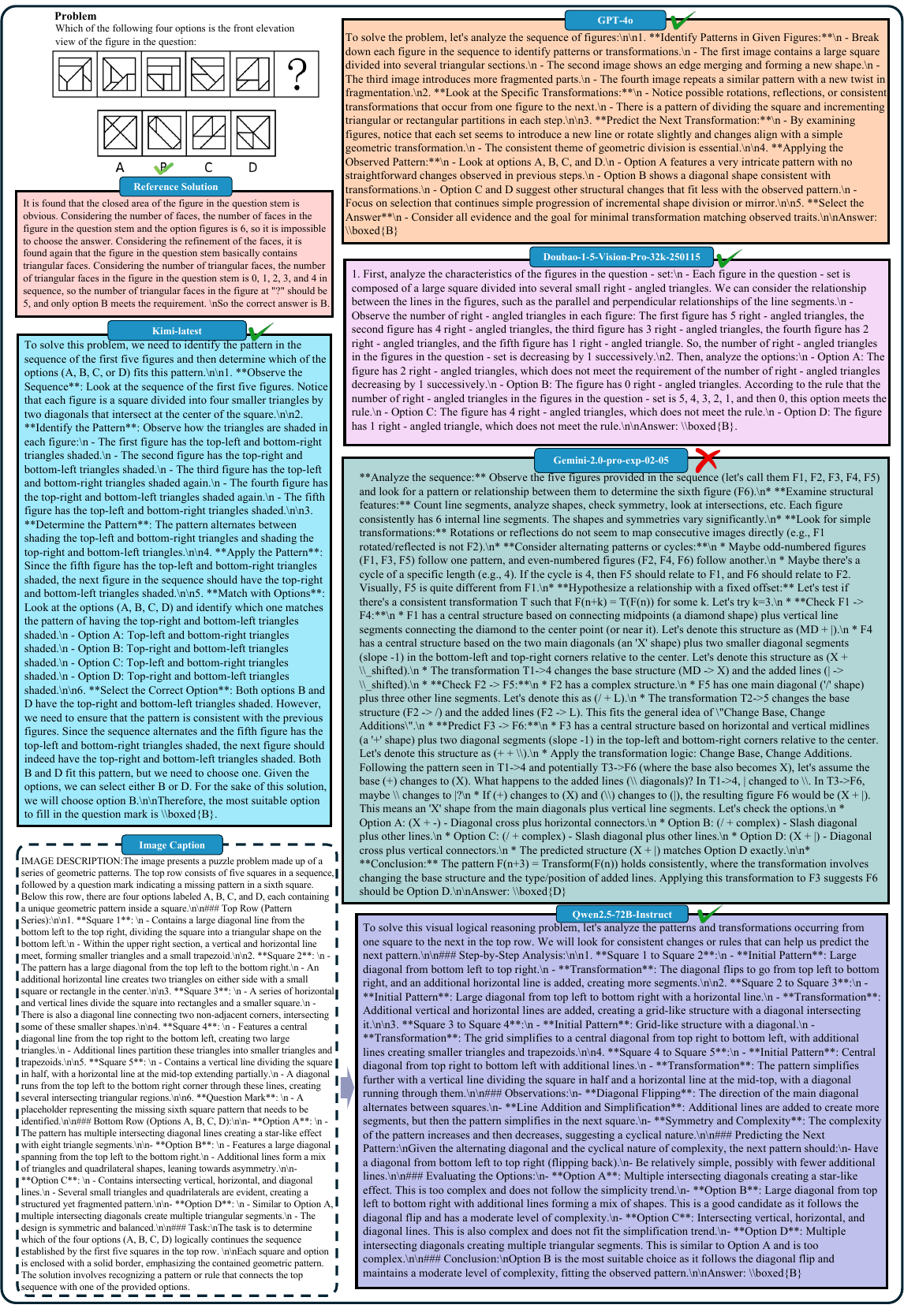}
    \caption{\textbf{Solution examples generated by different models.} Reference solution and outputs generated by GPT-4o, Kimi-latest, Gemini-2.0-pro-exp-02-05 and Doubao-1.5-Vision-Pro-32K. Additionally, the image caption and solution from LLMs (Qwen2.5-72B-Instruct) are also illustrated.}
    \label{fig:appendix_model_output_2}
\end{figure}

\begin{figure}
    \centering
    \includegraphics[width=1\linewidth]{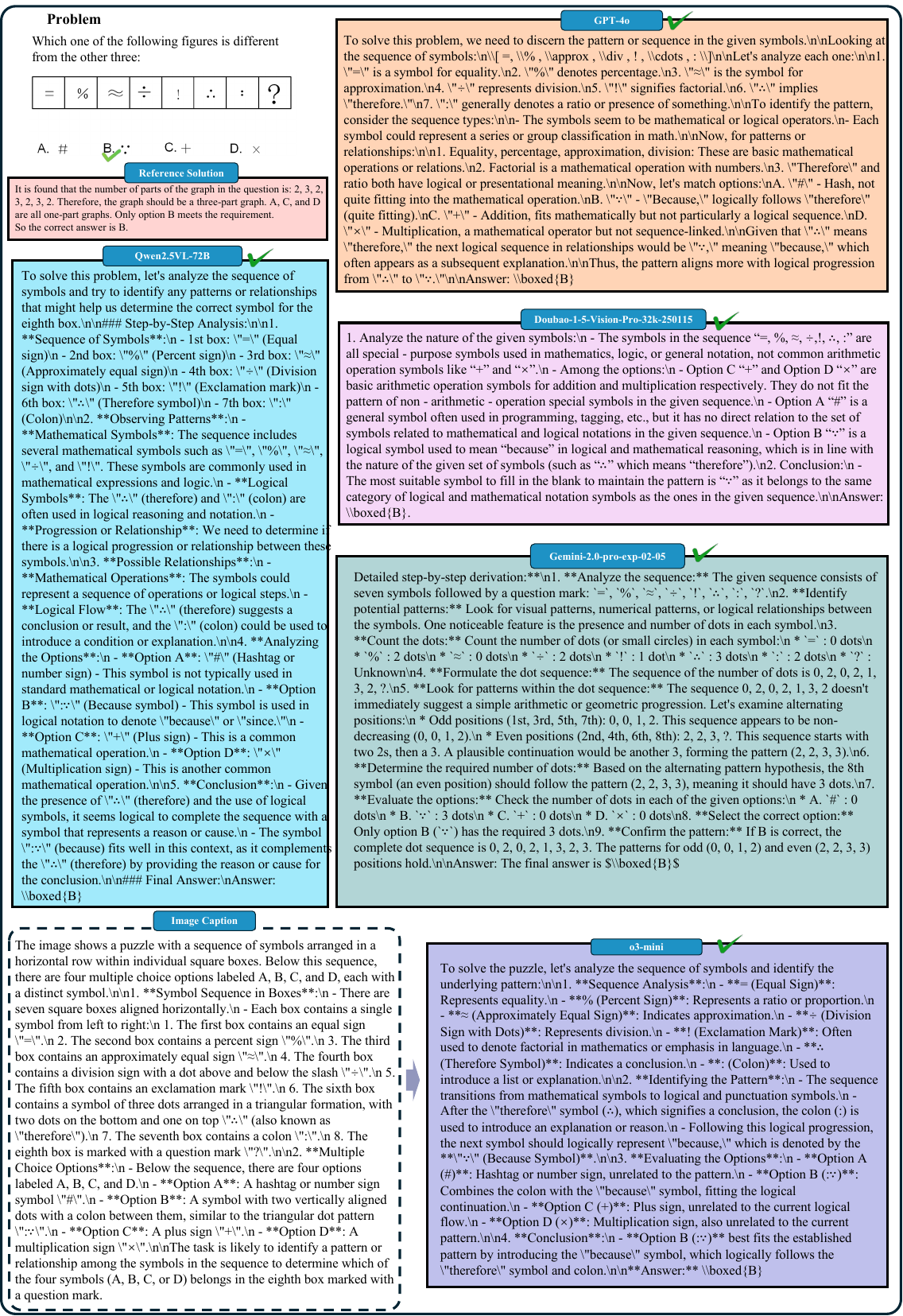}
    \caption{\textbf{Solution examples generated by different models.} Reference solution and outputs generated by GPT-4o, Qwen2.5VL-72B, Gemini-2.0-pro-exp-02-05 and Doubao-1.5-Vision-Pro-32k. Additionally, the image caption and solution from LLMs (o3-mini) are also illustrated.}
    \label{fig:appendix_model_output_3}
\end{figure}

We provide more solutions generated from different LLMs/MLLMs on our benchmark, as shown in Figure~\ref{fig:appendix_model_output_1}, Figure~\ref{fig:appendix_model_output_2} and Figure~\ref{fig:appendix_model_output_3}. For the majority of questions, almost all models fail to provide accurate solutions. Sometimes even when the final answer is correct, methodological wrong may persist.


\subsection{Hint Prompts Evaluation Details}
\begin{figure}[h]
     \centering
     \includegraphics[width=1.05\linewidth]{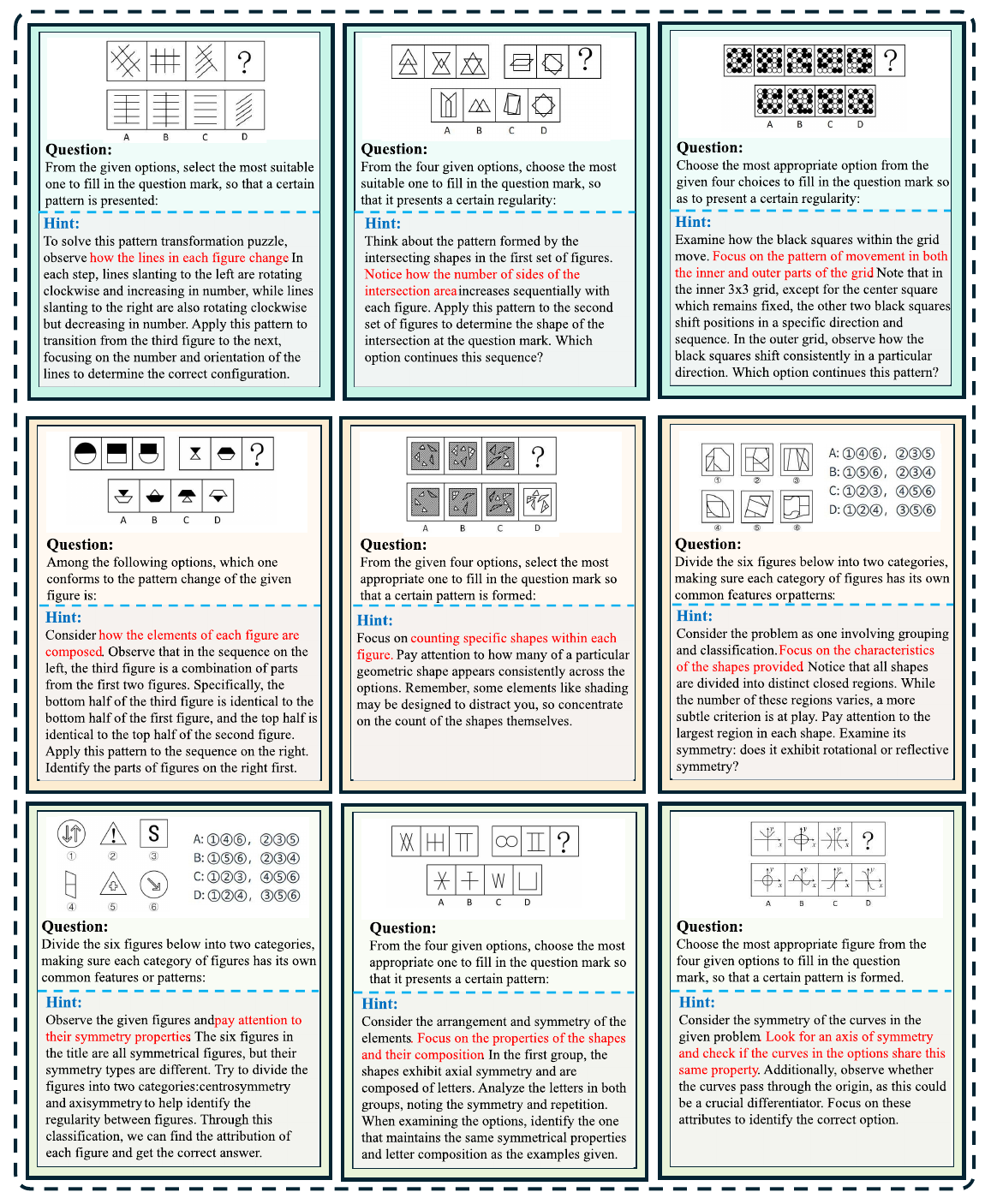}
     \caption{Examples of hint prompts. Hint prompts are provided to guide reasoning without revealing the final answer directly. }
     \label{fig:hint_examples}
 \end{figure}

We first generate hint prompts with GPT-4o, combining reference solutions with question data as inputs. All outputs undergo manual validation to prevent solution leakage. More examples are shown in Figure~\ref{fig:hint_examples}. After that, we input the hint prompts along with the same CoT prompt in CoT experiments (\enquote{\textit{Solve the complex visual logical reasoning problem through step-by-step reasoning. Think about the reasoning process first and answer the question following this format: Answer: \textbackslash boxed\{\$LETTER\}}.}) to MLLMs. 


\subsection{RL Experiments}
\noindent\textbf{Comparative SFT Experiments.}
To verify the effectiveness of RL method, we arrange the comparative SFT experiments on the same dataset as RL experiments. The instruction consists of questions and  Non-CoT prompts, and the responses are formatted direct answers.   

\noindent\textbf{RL Algorithm.} We employ  REINFORCE Leave-One-Out (RLOO)~\cite{ahmadian2024basicsrevisitingreinforcestyle} in our reinforcement learning training phase. As a critic-model-free algorithm,rloo is at a low computational cost while maintaining more robustness to noise and KL constraints.

\begin{samepage}
\noindent\textbf{Reward Modeling.}
Inspired by Deepseek-R1~\cite{deepseekai2025deepseekr1incentivizingreasoningcapability}, we design our rule-based reward system that mainly consists of two types of rewards:
\begin{enumerate}
\item {\textbf{Format rewards:}} To clarify model's outputs, we design a format rule that forces model to put its thinking process between ‘<think>’ and ‘</think>’ tags and put its final answer between ‘<answer>’ and ‘</answer>’ tags. Regular expression is applied to judge whether outputs conform to the format rule.
\item {\textbf{Accuracy rewards:}} The accuracy reward is decided by the response's correctness. The model should generate the response in right format, then the answer is extracted and judged whether it is matched to the correct option.
\end{enumerate}
\end{samepage}


\textbf{Hyperparameter settings.}
Our two RL models are trained with the hyperparameter configuration detailed in Table~\ref{hyperparameters}. And the hyperparameters used in SFT training stage are listed in Table~\ref{tab:hyperparameters_sft}.

\begin{table}[h]
    \centering
    \caption{Hyperparameter Settings for SFT Training Stage.}
    \renewcommand\arraystretch{1.5}
    \begin{tabular}{c | cc}
    \hline
        ~ & Qwen2.5-VL-7B-Instruct-SFT & InternVL2.5-38B-SFT  \\ \hline \hline
        pretrain model & Qwen2.5-VL-7B-Instruct & InternVL2.5-38B  \\ 
        learning rate & 0.5e-5 & 2e-5 \\ 
        batch size & 64 & 128 \\ 
        optimizer & AdamW & AdamW \\ 
        lr scheduler & cosine & cosine \\ 
        image strategy & image\_max\_pixels=262144 & max\_dynamic\_patch=6\\ 
        warmup ratio & 0.1 & 0.03 \\ 
        max epochs & 1 & 1  \\ 
        bf16 & True & True \\ \hline
    \end{tabular}
    \label{tab:hyperparameters_sft}
\end{table}
\begin{table}[h]
    \centering
    \caption{Hyperparameter Settings for RL Training Stage.}
    \renewcommand\arraystretch{1.5}
    \begin{tabular}{c | cc}
    \hline
        ~ & Qwen2.5-VL-7B-Instruct-RL & InternVL2.5-38B-RL  \\ \hline \hline
        pretrain model & Qwen2.5-VL-7B-Instruct & InternVL2.5-38B  \\ 
        RL Algorithm & rloo  & rloo  \\ 
       train batch size & 128 & 64  \\ 
        rollout batch\_size & 256 & 128  \\ 
        temperature & 1 & 1  \\ 
        n samples per prompt & 16 & 8  \\ 
        prompt max len & 1024 & 4096 \\ 
        generate max len & 3000 & 3000 \\ 
        bf16 & True & True \\ 
        actor learning rate & 1e-6 & 1e-6 \\ 
        init kl coef & 0 & 0 \\  \hline
    \end{tabular}
    \label{hyperparameters}
\end{table}
\textbf{Other Details.}
The training environment consists of CentOS Linux release 7.6.1810 operating system with CUDA 12.1. For Qwen2.5-VL-7B-Instruct-RL, we train for 80 steps on 1$\times$8 A800 GPUs and for InternVL2.5-38B-RL we train for 100 steps on 6$\times$8 A800 GPUs. 

\subsection{RL models Evaluation Details}
As mentioned above, we apply format rewards in RL experiments. Thus, to fully investigate the models' latent reasoning abilities, we utilize implement training-aligned prompts during evaluation in VisuLogic, which is shown as follows: \enquote{\textit{Solve the complex visual logical reasoning problem through step-by-step reasoning. Think about the reasoning process first and answer the question following this format: <think> THINKING </think><answer> ANSWER </answer>}}.

\subsection{Effectiveness of RL Experiments}
\begin{figure}[h]
     \centering
     \includegraphics[width=0.9\linewidth]{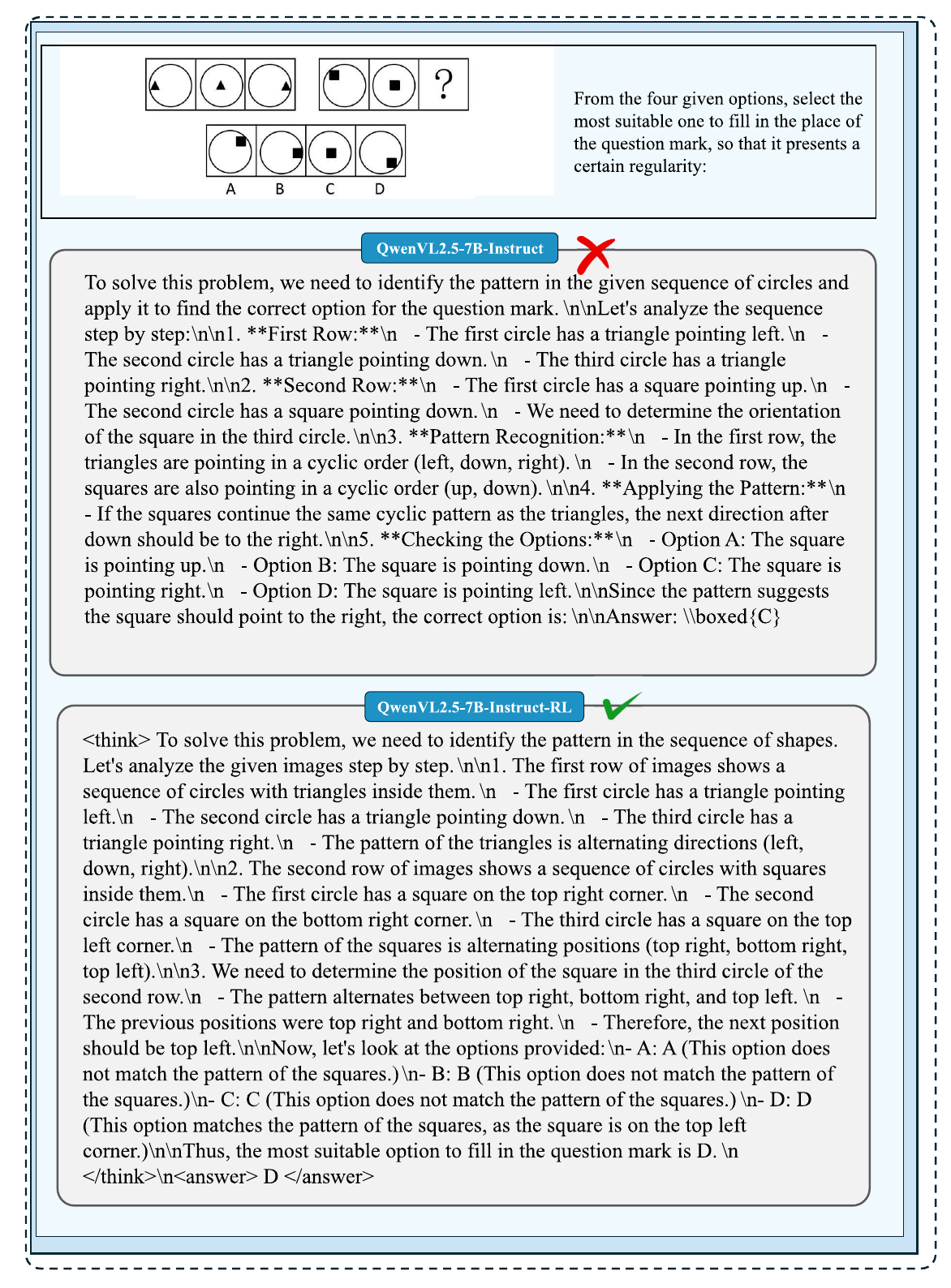}
     \caption{Comparison of model outputs before and after RL training stage for Qwen2.5-VL-7B. }
     \label{fig:RL_example_7b_1}
 \end{figure}

\begin{figure}[h]
    \centering
    \includegraphics[width=0.9\linewidth]{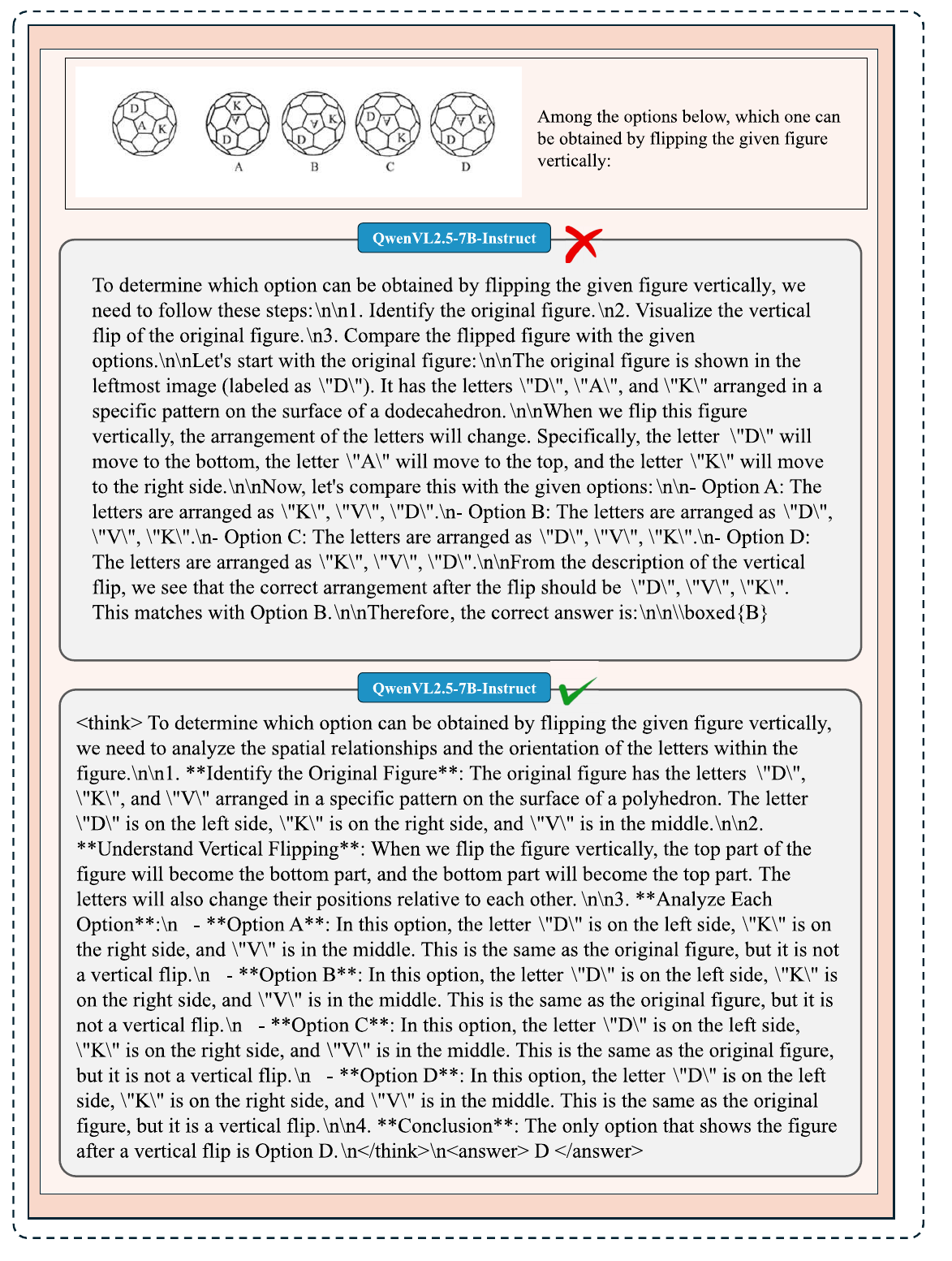}
    \caption{Comparison of model outputs before and after RL training stage for Qwen2.5-VL-7B.}
    \label{fig:RL_example_7b_2}
\end{figure}

\begin{figure}[h]
    \centering
    \includegraphics[width=0.9\linewidth]{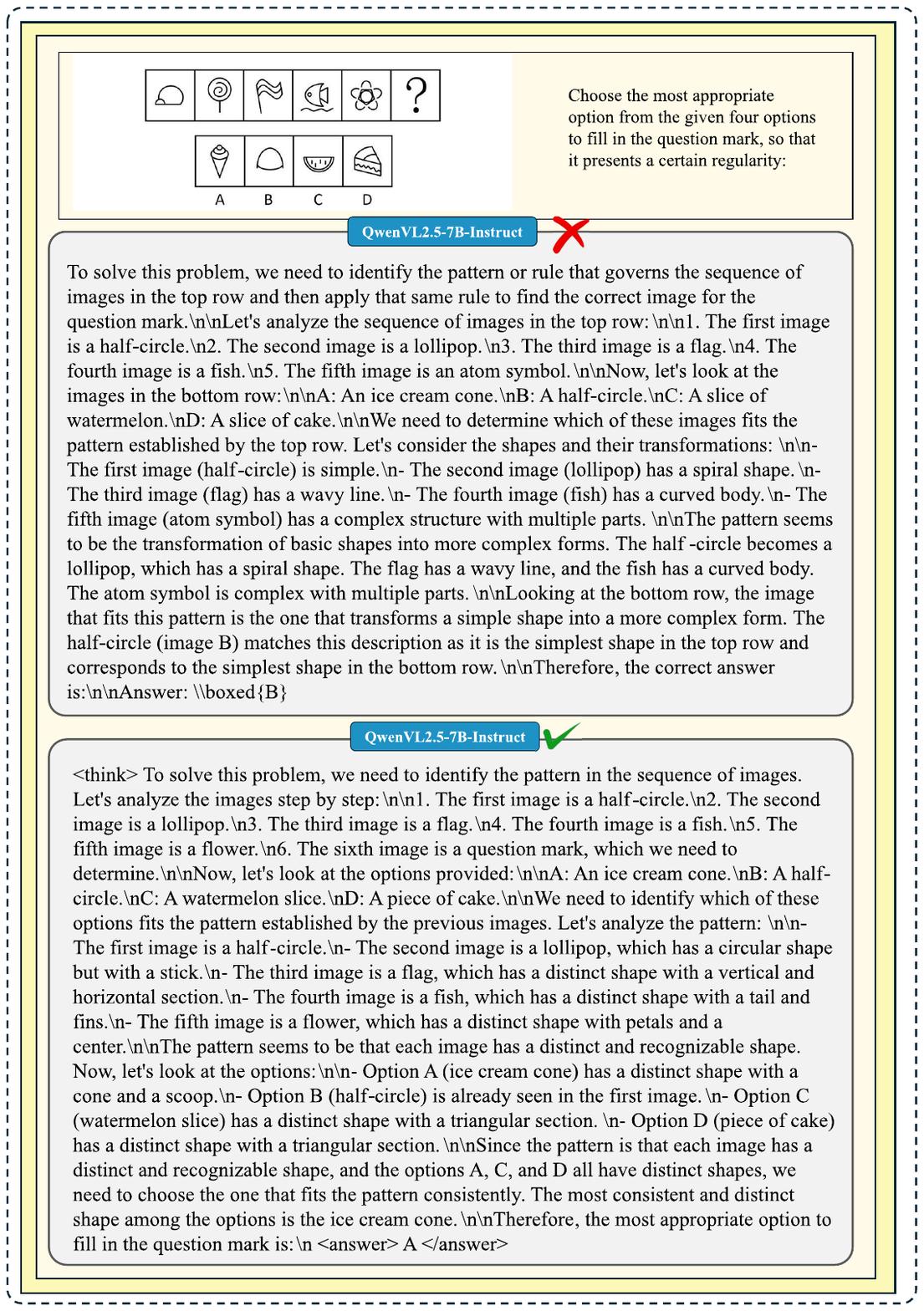}
    \caption{Comparison of model outputs before and after RL training stage for Qwen2.5-VL-7B.}
    \label{fig:RL_example_7b_3}
\end{figure}

\begin{figure}[h]
    \centering
    \includegraphics[width=0.9\linewidth]{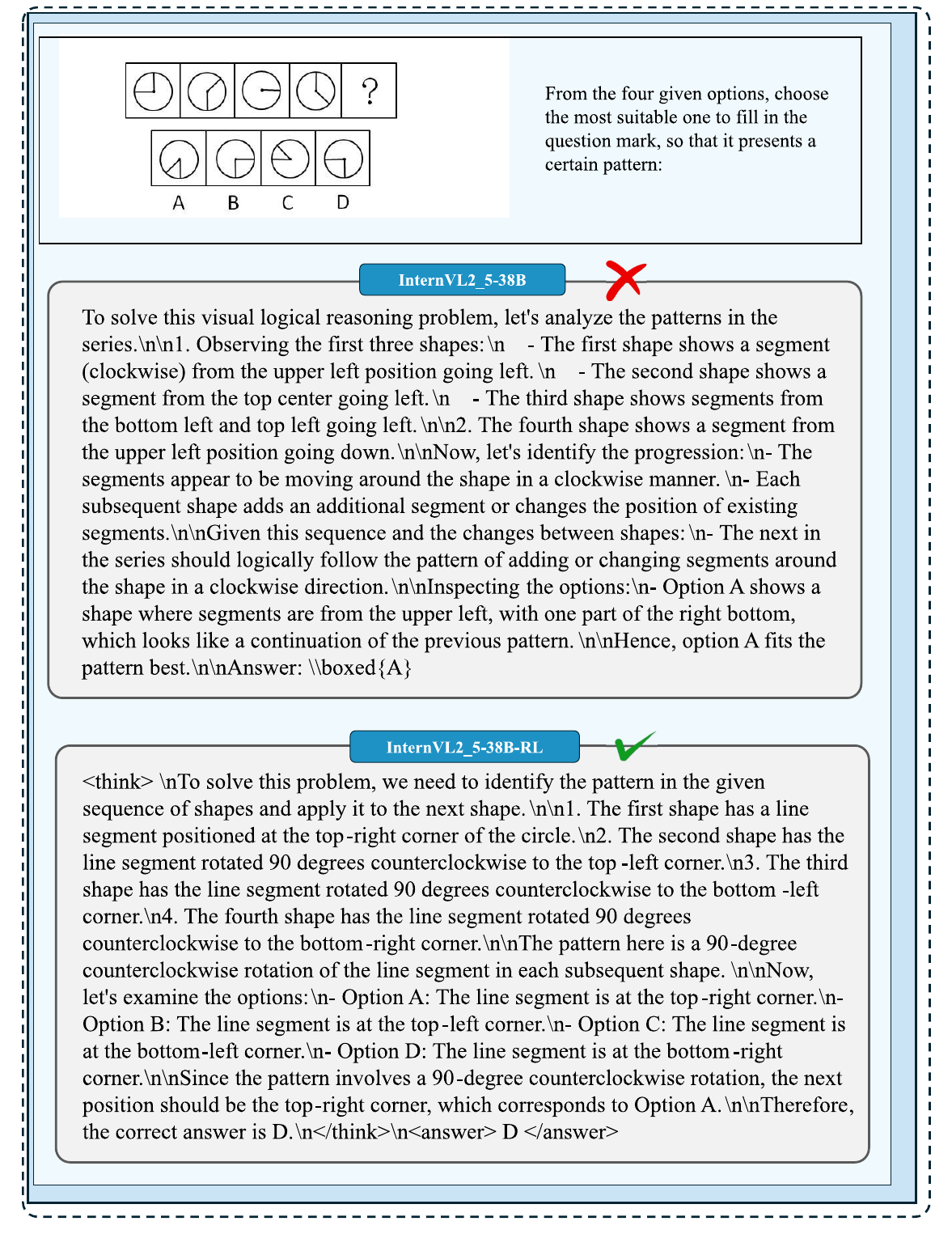}
    \caption{Comparison of model outputs before and after RL training stage for InternVL-2.5-38B.}
    \label{fig:RL_example_38b_1}
\end{figure}

\begin{figure}[h]
    \centering
    \includegraphics[width=0.9\linewidth]{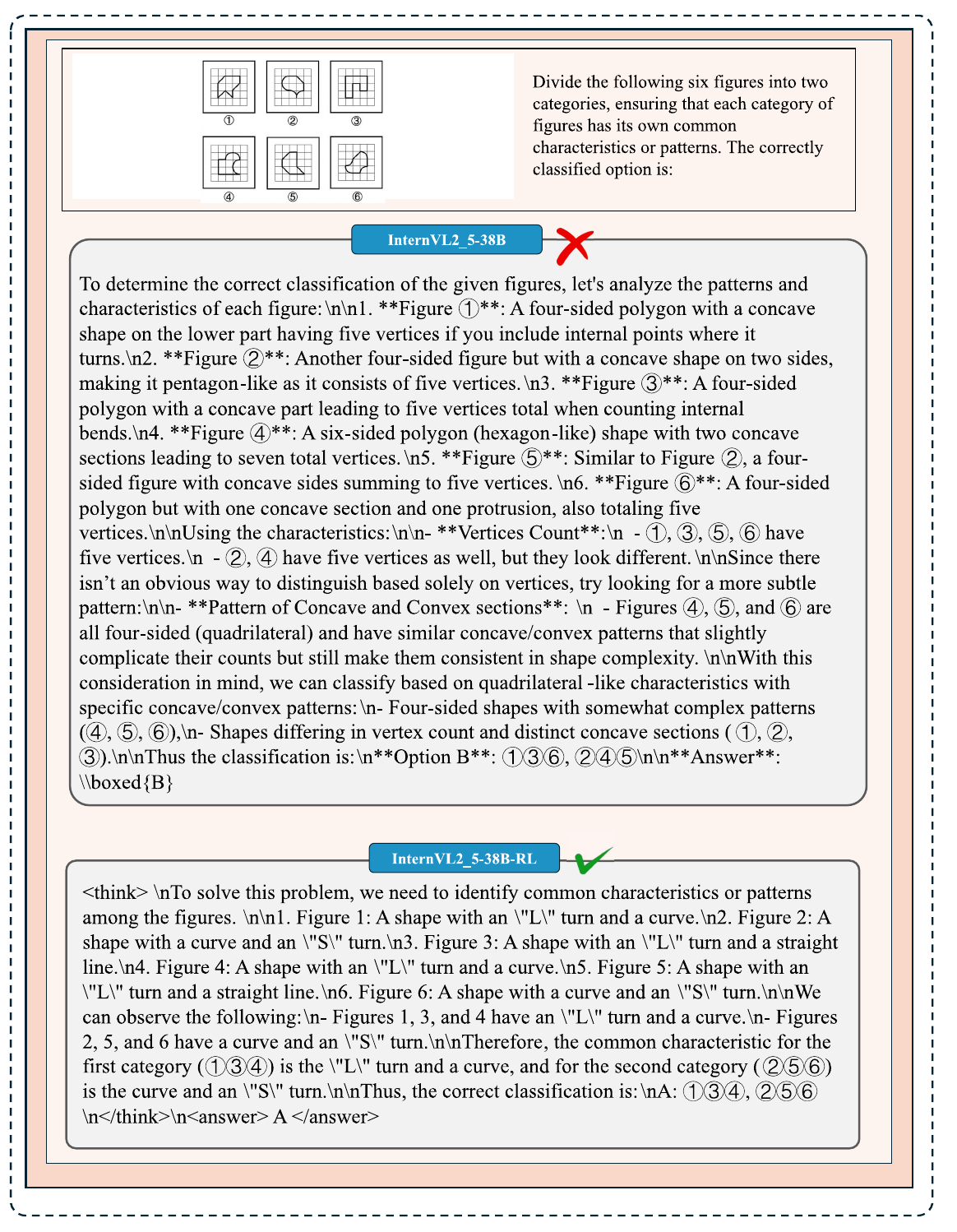}
    \caption{Comparison of model outputs before and after RL training stage for InternVL-2.5-38B.}
    \label{fig:RL_example_38b_2}
\end{figure}

\begin{figure}[h]
    \centering
    \includegraphics[width=0.9\linewidth]{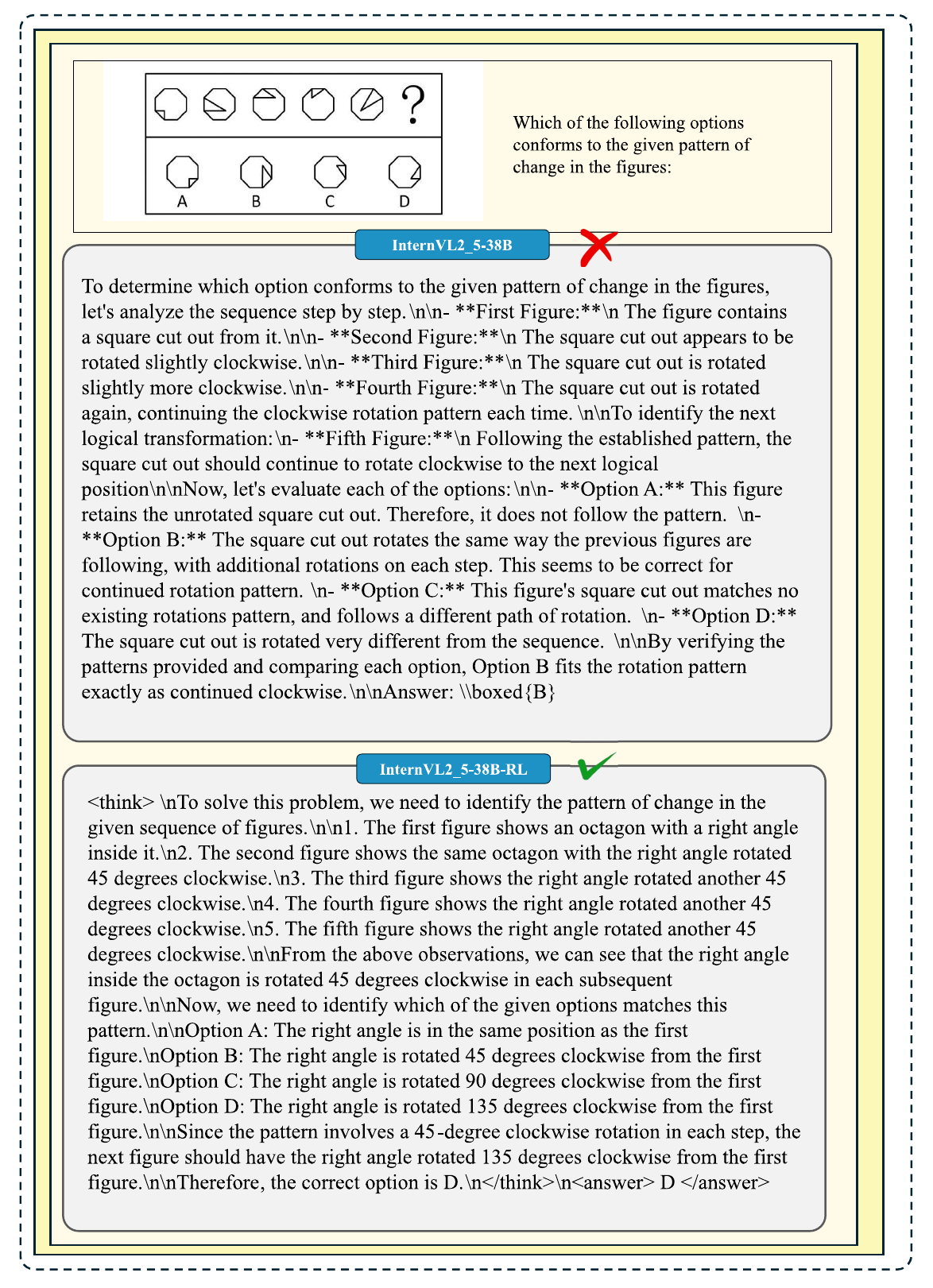}
    \caption{Comparison of model outputs before and after RL training stage for InternVL-2.5-38B.}
    \label{fig:RL_example_38b_3}
\end{figure}

Figures~\ref{fig:RL_example_7b_1}, \ref{fig:RL_example_7b_2}, \ref{fig:RL_example_7b_3}, \ref{fig:RL_example_38b_1}, \ref{fig:RL_example_38b_2} and \ref{fig:RL_example_38b_3}  demonstrate qualitative differences in model outputs between baseline and RL optimized models. It illustrates reinforcement learning (RL) training enables the model to perform fundamental-level analysis of reasoning tasks embedded in graphical representations.


\end{document}